\documentclass[reqno]{amsart}
\pdfoutput=1
\usepackage[english]{babel}
\usepackage{amsmath,amssymb,bm}
\usepackage{listings}
\usepackage{natbib}
\usepackage{graphicx}
\usepackage[linktocpage]{hyperref}

\newcommand{\E}{\mathsf{E}}
\newcommand{\VAR}{\mathsf{VAR}}

\newcommand{\Prob}{\mathsf{P}}

\newcommand{\expect}{\E}
\newcommand{\var}{\VAR}
\newcommand{\VarProb}{\mathsf{Q}}

\newcommand{\ud}{\textrm{d}}
\newcommand{\normaldist}{\mathcal{N}}
\newcommand{\gammadist}{\textrm{Gam}}
\newcommand{\unifdist}{\mathcal{U}}
\newcommand{\studenttdist}{\textrm{St}}
\newcommand{\const}{\textrm{const.}}
\newcommand{\trace}{\textrm{Tr}}
\newcommand{\varbound}{\mathcal{L}}
\newcommand{\data}{\mathcal{D}}
\newcommand{\proglang}[1]{\textsf{#1}}
\newcommand{\code}[1]{\lstinline!#1!}

\begin{document}

\title[VB inference for linear/logistic regression]{Variational Bayesian
inference\\for linear and logistic regression}
\author[Drugowitsch]{Jan Drugowitsch}
\address{Department of Neurobiology\\
Harvard Medical School\\
Boston, MA 02115\\
USA}
\email{jdrugo@gmail.com}

\begin{abstract}
  The article describe the model, derivation, and implementation of variational Bayesian inference for linear and logistic regression, both with and without automatic relevance determination.
  It has the dual function of acting as a tutorial for the derivation of variational Bayesian inference for simple models, as well as documenting, and providing brief examples for the \proglang{MATLAB}/\proglang{Octave} functions that implement this inference.
  These functions are freely available online.
\end{abstract}

\maketitle

\section{Introduction}

Linear and logistic regression are essential workhorses of statistical analysis, whose Bayesian treatment has received much recent attention \citep{gelman2013,bishop2006,murphy2012,hastie2011}.
These allow specifying the a-priori uncertainty and infer a-posteriori uncertainty about regression coefficients explicity and hierarchically, by, for example, specifying how uncertain we are a-priori that these coefficients are small.
However, Bayesian inference in such hierarchical models quickly becomes intractable, such that recent effort has focused on approximate inference, like Markov Chain Monte Carlo methods \citep{gilks1995}, or variational Bayesian approximation \citep{beal2003,bishop2006,murphy2012}.

Here, we describe such a variational treatment and implementation of Bayesian hierarchical models for both linear and logistic regression.
Even though neither the statistical models nor their Bayesian approximation are particularly novel, the article provides a tutorial-style introduction to the derivation of their algorithms, together with a \proglang{MATLAB}/\proglang{Octave} implementation of these algorithms.
As such, it bridges the gap between theory and practice of derivation and implementation.

The presentation of the variational inference derivation is closely aligned to that of \cite{bishop2006}, but with essential differences.
Specifically, both models include a variant with automatic relevance determination (ARD), which consists of assigning an individual hyper-prior to each regression coefficient separately.
These hyper-priors are adjusted to eventually prune irrelevant coefficients \citep{wipf2007} without the need for a separate validation set, unlike comparable sparsity-inducing methods like the Lasso \citep{tibshirani1996}.
\cite{bishop2006} describes ARD only in the context of type-II maximum likelihood \citep{mackay1992,neal1996,tipping2001}, where it (hyper-)parameters are tuned by maximizing the marginal likelihood (or \emph{model evidence}).
Here, instead, we apply the full Bayesian treatment, and find the ARD hyper-posteriors by variational Bayesian inference.

The model underlying linear regression is closely related to \cite{griffin2010}, where the authors analyze the influence of prior choice on regression coefficient shrinkage.
They promote priors in the form of scale mixtures of zero-mean normals, which allow for a larger difference in regression coefficients than would be possible under a normal prior.
The authors proceed by suggesting a normal-gamma prior and analyze its shrinkage properties in detail.
The prior used in this work is also member of the scale mixtures of normals, and thus shares its advantageous properties.
However, instead of a normal-gamma prior, this work uses a normal inverse-gamma prior in combination with another inverse-gamma hyper-prior, as its conjugacy to the likelihood is advantageous for use with variational Bayesian inference.
The same analysis performed by \cite{griffin2010} for the normal-gamma case should be amendable to the normal inverse-gamma case, but has yet to be performed.

The article is structured as follows.
It first described linear regression, followed by logistic regression.
For each of these, it first introduces the generative model, followed by deriving the variational Bayesian approximation.
After that it introduces the ARD variants, together with their required changes to variational inference.
This is followed by a detailed description of the \proglang{MATLAB}/\proglang{Octave} functions that implement this inference, and a set of examples that demonstrate their use.

All functions are implemented in the \texttt{VBLinLogit} library, which can be found at \url{https://github.com/DrugowitschLab/VBLinLogit}.
The line numbers in this paper refer to \texttt{v0.3} of this libary.

\section{Linear regression}

This section describes inference in a model performing linear regression.
It is similar to that in \cite{bishop2006} by assuming a hyper-prior $\alpha$ on the regression coefficients $\bm{w}$.
However, rather than just inferring the posterior $\bm{w}$, as done in
\cite{bishop2006}, it additionally puts an inverse-gamma prior on the variance
$\tau^{-1}$ and infers the joint posterior of $\bm{w}$ and $\tau$ jointly.
Furthermore, \cite{bishop2006} utilizes type-II maximum likelihood to deal with
automatic relevance determination for linear regression.
Here, we use the variational Bayesian approximation instead.

\subsection{The model}

The model assumes a linear relation between $D$-dimensional inputs
$\bm{x}$ and outputs $y$ and constant-variance Gaussian noise, such
that the data likelihood is given by
\begin{equation}
  \Prob(y | \bm{x}, \bm{w}, \tau) = \normaldist(y | \bm{w}^\top \bm{x},
  \tau^{-1}) = \left( \frac{\tau}{2 \pi} \right)^{1/2} \exp \left( -
    \frac{\tau}{2} ( y - \bm{w}^\top \bm{x})^2 \right) .
\end{equation}
Given all data $\data = \{ \bm{X}, \bm{Y} \}$, with $\bm{X} = \{
\bm{x}_1, \dots, \bm{x}_N \}$ and $\bm{Y} = \{ y_1, \dots, y_N \}$,
the data likelihood is
\begin{equation}
  \Prob(\bm{Y} | \bm{X}, \bm{w}, \tau) = \prod_n \Prob(y_n | \bm{x}_n, \bm{w},
  \tau) .
\end{equation}
The prior on $\bm{w}$ and $\tau$ is conjugate normal inverse-gamma
\begin{eqnarray}
  \Prob(\bm{w}, \tau | \alpha) &=& \normaldist(\bm{w}| 0, (\tau \alpha)^{-1}
  \bm{I}) \gammadist(\tau | a_0, b_0) \nonumber \\
  &=& \left( \frac{\alpha}{2 \pi} \right)^{D/2}
  \frac{b_0^{a_0}}{\Gamma(a_0)} \tau^{D/2 + a_0 - 1} \exp \left( -
    \frac{\tau}{2} (\alpha \bm{w}^\top \bm{w} + 2 b_0) \right) ,
\end{eqnarray}
parametrized by $\alpha$.
As in \cite{griffin2010}, this prior is member of the scale mixtures of normals.
In this prior, $\tau$ appears as $\tau^{-1}$ in the variance of the zero-mean normal on $\bm{w}$.
Due to the gamma on $\tau$, this $\tau^{-1}$ is inverse-gamma with shape $a_0$, scale $b_0$, and moments $\expect \left( \tau^{-1} \right) = b_0 / (a_0 - 1)$ for $a_0 > 1$ and $\var \left( \tau^{-1} \right) = b_0^2 / \left( (a_0 - 1)^2 (a_0 - 2) \right)$ for $a_0 > 2$.
The hyper-parameter $\alpha$ is assigned the hyper-prior
\begin{equation}
  \Prob( \alpha ) = \gammadist( \alpha | c_0, d_0 ) = \frac{1}{\Gamma(c_0)} d_0^{c_0}
  \alpha^{c_0 - 1} \exp(- d_0 \alpha) ,
\end{equation}
with moments of $\alpha^{-1}$ analogous to $\tau^{-1}$.
Due to the hyper-prior, there is no analytic solution to the
posteriors, and variational Bayesian inference will be applied.

\subsection{Variational Bayesian inference}

The variational posteriors are found by maximizing the variational bound
\begin{equation}
  \varbound(\VarProb) = \iiint \VarProb(\bm{w}, \tau, \alpha) \ln \frac{\Prob(\bm{Y} |
    \bm{X}, \bm{w}, \tau) \Prob(\bm{w}, \tau | \alpha) \Prob(\alpha)}{\VarProb(\bm{w},
    \tau, \alpha)} \ud \bm{w} \ud \tau \ud \alpha \le \ln \Prob(\data) ,
\end{equation}
where $\Prob(\data)$ is the model evidence.
To maximize this bound, we assume that the variational distribution $\VarProb(\bm{w}, \tau, \alpha)$, which approximates the posterior $\Prob(\bm{w}, \tau, \alpha | \data)$, factors into $\VarProb(\bm{w}, \tau) \VarProb(\alpha)$.

Using standard results from variational Bayesian inference \citep{beal2003,bishop2006}, the variational posterior for $\bm{w}, \tau$ that maximizes the variational bound $\varbound(\VarProb)$ while holding $\VarProb(\alpha)$ fixed, is given by
\begin{eqnarray}
  \ln \VarProb^*(\bm{w}, \tau) &=& \ln \Prob(\bm{Y} | \bm{X}, \bm{w}, \tau) +
  \expect_\alpha(\ln \Prob(\bm{w}, \tau | \alpha)) + \const \nonumber \\
  &=& \left( \frac{D}{2} + a_0 - 1 + \frac{N}{2} \right) \ln \tau
  \nonumber \\
  && \quad - \frac{\tau}{2} \Bigg( \bm{w}^\top \left( \expect_\alpha(\alpha)
    \bm{I} + \sum_n \bm{x}_n \bm{x}_n^\top \right) \bm{w}\nonumber \\
  && \qquad \qquad + \sum_n y_n^2
  - 2 \bm{w}^\top \sum_n \bm{x}_n y_n + 2 b_0 \Bigg) + \const \nonumber \\
  &=& \ln \normaldist(\bm{w} | \bm{w}_N, \tau^{-1} \bm{V}_N)
  \gammadist(\tau | a_N, b_N) ,
  \label{eq:vb_linear_w_tau_post}
\end{eqnarray}
with
\begin{eqnarray}
  \bm{V}_N^{-1} &=& \expect_\alpha(\alpha) \bm{I} + \sum_n \bm{x}_n
  \bm{x}_n^\top , \\
  \bm{w}_N &=& \bm{V}_N \sum_n \bm{x}_n y_n , \\
  a_N &=& a_0 + \frac{N}{2}, \label{eq:vblin_an_update} \\
  b_N &=& b_0 + \frac{1}{2} \left( \sum_n y_n^2 - \bm{w}_N^\top
    \bm{V}_N^{-1} \bm{w}_N \right) \nonumber \\
  &=& b_0 + \frac{1}{2} \left( \sum_n (
    y_n - \bm{w}_N^\top \bm{x}_n )^2 + \expect_\alpha(\alpha) \bm{w}_N^\top
    \bm{w}_N \right) .
\end{eqnarray}
The variational posterior for $\alpha$ is
\begin{eqnarray}
  \ln \VarProb^*(\alpha) &=& \expect_{\bm{w}, \tau}(\ln \Prob(\bm{w}, \tau |
  \alpha)) + \ln \Prob(\alpha) + \const \nonumber \\
  &=& \left( c_0 - 1 + \frac{D}{2} \right) \ln \alpha - \alpha \left(
    d_0 + \frac{1}{2} \expect_{\bm{w}, \tau}(\tau \bm{w}^\top \bm{w})
  \right) + \const \nonumber \\
  &=& \ln \gammadist(\alpha | c_N, d_N) ,
\end{eqnarray}
with
\begin{eqnarray}
  c_N &=& c_0 + \frac{D}{2} , \\
  d_N &=& d_0 + \frac{1}{2} \expect_{\bm{w}, \tau} (\tau \bm{w}^\top
  \bm{w}) .
\end{eqnarray}
The expectations are evaluated with respect to the variational
posterior and are given by
\begin{eqnarray}
  \expect_{\bm{w}, \tau}(\tau \bm{w}^\top \bm{w}) &=& \frac{a_N}{b_N}
  \bm{w}_N^\top \bm{w}_N + \trace(\bm{V}_N) , \\
  \expect_\alpha(\alpha) &=& \frac{c_N}{d_N} .
\end{eqnarray}

The variational bound itself consists of
\begin{eqnarray}
  \varbound(\VarProb) &=& \expect_{\bm{w}, \tau}(\ln \Prob(\bm{Y} | \bm{X},
  \bm{w}, \tau)) + \expect_{\bm{w}, \tau, \alpha}(\ln \Prob(\bm{w}, \tau |
  \alpha)) \nonumber \\
  && \quad + \expect_\alpha(\ln \Prob(\alpha)) - \expect_{\bm{w}, \tau}(\ln
  \VarProb(\bm{w}, \tau)) \nonumber \\
  && \quad - \expect_\alpha(\ln \VarProb(\alpha)) , \\
  \expect_{\bm{w}, \tau}( \ln \Prob(\bm{Y} | \bm{X}, \bm{w}, \tau)) &=&
  \frac{N}{2} \left( \psi(a_N) - \ln b_N - \ln 2 \pi \right) \nonumber
  \\
  && \quad - \frac{1}{2} \sum_n \left( \frac{a_N}{b_N} (y_n -
    \bm{w}_N^\top \bm{x}_n)^2 + \bm{x}_n^\top \bm{V}_N \bm{x}_n \right) , \\
  \expect_{\bm{w}, \tau, \alpha}(\ln \Prob(\bm{w}, \tau | \alpha)) &=&
  \frac{D}{2} \left( \psi(a_N) - \ln b_N + \psi(c_N) - \ln d_N - \ln 2
    \pi \right) \nonumber \\
  && \quad - \frac{1}{2} \frac{c_N}{d_N} \left( \frac{a_N}{b_N}
    \bm{w}_N^\top \bm{w}_N + \trace(\bm{V}_N) \right) \nonumber \\
  && \quad - \ln \Gamma(a_0) + a_0 \ln b_0 \nonumber \\
  && \quad + (a_0 - 1) (\psi(a_N) - \ln b_N) - b_0 \frac{a_N}{b_N} , \\
  \expect_\alpha(\ln \Prob(\alpha)) &=& - \ln \Gamma(c_0) + d_0 \ln c_0 \nonumber \\
  && \quad + (c_0 - 1) (\psi(c_N) - \ln d_N) - d_0 \frac{c_N}{d_N} , \\
  \expect_{\bm{w}, \tau}(\ln \VarProb(\bm{w}, \tau)) &=& \frac{D}{2} (
  \psi(a_N) - \ln b_N - \ln 2 \pi - 1) - \frac{1}{2} \ln | \bm{V}_N |
  \nonumber \\
  && \quad - \ln \Gamma(a_N) + a_N \ln b_N \nonumber \\
  && \quad + (a_N - 1)(\psi(a_N) -
  \ln b_N) - a_N, \\
  \expect_\alpha(\ln \VarProb(\alpha)) &=& -\ln \Gamma(c_N) + (c_N - 1)
  \psi(c_N) + \ln d_N - c_N .
\end{eqnarray}
In combination, this gives
\begin{eqnarray}
  \varbound(\VarProb) &=& - \frac{N}{2} \ln 2 \pi - \frac{1}{2} \sum_n \left(
    \frac{a_N}{b_N} (y_n - \bm{w}_N^\top \bm{x}_n)^2 + \bm{x}_n^\top
    \bm{V}_N \bm{x}_n \right) + \frac{1}{2} \ln | \bm{V}_N | +
  \frac{D}{2} \nonumber \\
  && \quad - \ln \Gamma(a_0) + a_0 \ln b_0 - b_0 \frac{a_N}{b_N} +
  \ln \Gamma(a_N) - a_N \ln b_N + a_N \nonumber \\
  && \quad - \ln \Gamma(c_0) + c_0 \ln d_0  + \ln \Gamma(c_N) - c_N \ln d_N
  \label{eq:vb_linear_var_bound}
\end{eqnarray}
This bound is maximized by iterating over the updates for $\bm{V}_N$,
$\bm{w}_N$, $a_N$, $b_N$, $c_N$, and $d_N$ until $\varbound(\VarProb)$
reaches a plateau.

\subsection{Predictive density}

The predictive density is evaluated by approximating the posterior
$\Prob(\bm{w}, \tau | \data)$ by its variational counterpart $\VarProb(\bm{w},
\tau)$, to get
\begin{eqnarray}
  \Prob(y | \bm{x}, \data) &=& \iint \Prob(y | \bm{x}, \bm{w}, \tau) \Prob(\bm{w},
  \tau | \data) \ud \bm{w} \ud \tau \nonumber \\
  &\approx& \iint \Prob(y | \bm{x}, \bm{w}, \tau) \VarProb(\bm{w}, \tau) \ud
  \bm{w} \ud \tau \nonumber  \\
  &=& \iint \normaldist( y | \bm{w}^\top \bm{x}, \tau^{-1} ) \normaldist(
  \bm{w} | \bm{w}_N, \tau^{-1} \bm{V}_N) \gammadist( \tau | a_N, b_N )
  \ud \bm{w} \ud \tau \nonumber \\
  &=& \int \normaldist( y | \bm{w}_N^\top \bm{x}, \tau^{-1} (1 + \bm{x}^\top
  \bm{V}_N \bm{x})) \gammadist(\tau | a_N, b_N) \ud \tau \nonumber \\
  &=& \studenttdist \left(y | \bm{w}_N^\top \bm{x}, (1 + \bm{x}^\top \bm{V}_N
  \bm{x})^{-1} \frac{a_N}{b_N}, 2 a_N \right) ,
  \label{eq:vb_linear_pred}
\end{eqnarray}
where standard results of convolving Gaussians with other Gaussians
and Gamma distributions where used \citep{bishop2006,murphy2012}.
The resulting distribution is a
Student's t distribution with mean $\bm{w}_N^\top \bm{x}$, precision $(1 +
\bm{x}^\top \bm{V}_N \bm{x})^{-1} a_N / b_N$, and $2 a_N$ degrees of
freedom. The resulting predictive variance is $(1 + \bm{x}^\top \bm{V}_N \bm{x})
b_N / (a_N - 1)$.

\subsection{Using automatic relevance determination}

Automatic relevance determination (ARD) determines the relevance of
the elements of the input to determine the output by assigning a
separate shrinkage prior to each element of the weight vector, which
is in turn adjusted by a hyper-prior. While the data likelihood
remains unchanged, the prior on $\bm{w}, \tau$ is modified to be
\begin{eqnarray}
  \Prob(\bm{w}, \tau | \bm{\alpha}) &=& \normaldist( \bm{w} | \bm{0},
  (\tau \bm{A})^{-1}) \gammadist(\tau | a_0, b_0) \nonumber \\
  &=& \frac{|\bm{A}|^{1/2}}{\sqrt{2 \pi}^D}
  \frac{b_0^{a_0}}{\Gamma(a_0)} \tau^{D/2 + a_0 - 1} \exp \left( -
    \frac{\tau}{2} ( \bm{w}^\top \bm{A} \bm{w} + 2 b_0 ) \right) ,
\end{eqnarray}
where the vector $\bm{\alpha} = ( \alpha_1, \dots, \alpha_D)^\top$ forms
the diagonal of $\bm{A}$. All of the $\alpha$'s are independent, such
that the hyper-prior is given by
\begin{equation}
  \Prob(\bm{\alpha}) = \prod_i \gammadist(\alpha_i | c_0, d_0) = \prod_i
  \frac{1}{\Gamma(c_0)} d_0^{c_0} \alpha_i^{c_0 - 1} \exp( - d_0
  \alpha_i ) .
\end{equation}

Variational Bayesian inference is performed as before, resulting in the
variational posteriors
\begin{equation}
  \VarProb^*(\bm{w}, \tau) = \normaldist(\bm{w} | \bm{w}_N, \tau^{-1}
\bm{V}_N) \gammadist(\tau | a_N, b_N), \quad \VarProb^*(\bm{\alpha}) =
\prod_i \gammadist(\alpha_i | c_N, d_{Ni}) ,
\end{equation}
with parameters
\begin{eqnarray}
  \bm{V}_N^{-1} &=& \expect_\alpha(\bm{A}) + \sum_n \bm{x}_n
  \bm{x}_n^\top, \\
  \bm{w}_N &=& \bm{V}_N \sum_n \bm{x}_n y_n , \\
  a_N &=& a_0 + \frac{N}{2} \\
  b_N &=& b_0 + \frac{1}{2} \left( \sum_n y_n^2 - \bm{w}_N^\top
    \bm{V}_N^{-1} \bm{w}_N \right) \nonumber \\
  &=& b_0 + \frac{1}{2} \left( \sum_n (\bm{w}_N^\top \bm{x}_N - y_n)^2 +
    \bm{w}_N^\top \expect_\alpha(\bm{A}) \bm{w}_N \right), \\
  c_N &=& c_0 + \frac{1}{2} , \\
  d_{Ni} &=& d_0 + \frac{1}{2} \expect_{\bm{w}, \tau}(\tau w_i^2) ,
\end{eqnarray}
with expectations $\expect_{\bm{w}, \tau}(\tau w_i^2) = w_{Ni}^2 a_N
/ b_N + (\bm{V}_N)_{ii}$, and $\expect_\alpha(\bm{A}) = \bm{A}_N$ is a
diagonal matrix with elements $\expect_\alpha(\alpha_i) = c_N /
d_{Ni}$.

The variational bound changes to
\begin{eqnarray}
  \varbound(\VarProb) &=& - \frac{N}{2} \ln 2 \pi - \frac{1}{2} \sum_n \left(
    \frac{a_N}{b_N} (y_n - \bm{w}_N^\top \bm{x}_n)^2 + \bm{x}_n^\top
    \bm{V}_N \bm{x}_n \right) + \frac{1}{2} \ln | \bm{V}_N | +
  \frac{D}{2} \nonumber \\
  && \quad - \ln \Gamma(a_0) + a_0 \ln b_0 - b_0 \frac{a_N}{b_N} +
  \ln \Gamma(a_N) - a_N \ln b_N + a_N \nonumber \\
  && \quad + \sum_i \left( - \ln \Gamma(c_0) + c_0 \ln d_0  + \ln
    \Gamma(c_N) - c_N \ln d_{Ni} \right) .  
\end{eqnarray}
The predictive distribution remains unchanged, as the prior does not
appear in the expression for the variational posterior $\Prob(\bm{w},
\tau)$.

\subsection{Implementation}
\label{sec:vblin_implementation}

The scripts that compute the variational posterior parameters without and with ARD
are \code{vb\_linear\_fit.m} and \code{vb\_linear\_fit\_ard.m}, respectively.
\code{vb\_linear\_pred.m} computes the predictive density parameters for a set of input vectors.

\subsubsection{Variational posterior parameters without ARD}

The function \code{vb\_linear\_fit.m} computes the variational posterior parameters without ARD, and has syntax
\begin{lstlisting}
[w, V, invV, logdetV, an, bn, E_a, L] =
    vb_linear_fit(X, y[, a0, b0, c0, d0])
\end{lstlisting}
where \code{X} is the $N \times D$ input matrix with $\bm{x}_n^\top$ as its rows, and \code{y} is a column vector, containing all $y_n$'s.
The optional parameters, \code{a0}, \code{b0}, \code{c0}, and \code{d0}, specify the prior parameters $a_0$, $b_0$, $c_0$, and $d_0$, respectively.
If not given, they default to $a_0 = 10^{-2}$, $b_0 = 10^{-4}$, $c_0 = 10^{-2}$, and $d_0 = 10^{-4}$.
For these values, the mean of $\tau^{-1}$ is undefined, but its mode is at $b_0 / (a_0 + 1) \approx 10^{-4}$, implying the a-prior most likely variance of the prior on $\bm{w}$ to be small.
The variance of $\tau^{-1}$ is also undefined for $a_0 \le 2$, but the related variance on $\tau$ is $\var(\tau) = a_0 / b_0^2 = 10^6$.
Thus, even though the default prior on $\tau$ implies some shrinkage of $\bm{w} \to \bm{0}$, this shrinkage is weak due to the prior's large width.
Furthermore, the update Eq.~(\ref{eq:vblin_an_update}) of $a_N$ reveals that $a_0$ can be interpreted as the half the a-prior number of observations.
Thus the prior has the weight of $2 \times 10^{-2}$ observations, and thus loses its influence with very few observations.
The same applies for the prior on $\alpha$.

The returned values, \code{w}, \code{V}, \code{an}, and \code{bn} correspond to the normal inverse-gamma parameters $\bm{w}_N$, $\bm{V}_N$, $a_N$, and $b_N$, respectively, of the variational posterior $\VarProb^*(\bm{w},\tau)$, Eq.~(\ref{eq:vb_linear_w_tau_post}).
The variational parameters for $\VarProb^*(\alpha)$ are summarized by the returned $\textrm{\code{E\_a}} = \expect_\alpha(\alpha)$.
The function additionally returns $\textrm{\code{invV}} = \bm{V}_N^{-1}$, and $\textrm{\code{logdetV}} = \ln | \bm{V}_N |$, such that, if required, these values do not need to be re-computed.
The returned \code{L} is the variational bound $\varbound(\VarProb)$, Eq.~(\ref{eq:vb_linear_var_bound}), evaluated at the returned parameters.

After initializing the required data structures, initializing parameters, and pre-computing some constants, the function finds the variational posterior parameters by updating $\VarProb^*(\bm{w}, \tau)$ (lines 65--73) and $\VarProb^*(\alpha)$ (lines 75--77).
After each update, it evaluates the parameter-dependent components of the variational bound $\varbound(\VarProb)$ in lines 79--82.
This is repeated until either the change in $\varbound(\VarProb)$ between two consecutive iterations drops below $0.001\%$, or the number of iterations exceeds 500.

To update $\bm{V}_N^{-1}$, $\bm{w}_N$, $b_n$, and $\varbound(\VarProb)$, the function vectorizes operations over $n$ by
\begin{eqnarray}
  \sum_n \bm{x}_n \bm{x}_n^\top & = & \textrm{\code{X' * X}}, \\
  \sum_n \bm{x}_n y_n & = & \textrm{\code{X' * y}}, \\
  \bm{w}_N^\top \bm{x}_n & = & \left( \textrm{\code{X * w}} \right)_n, \\
  \sum_n \bm{x}_n^\top \bm{V}_N \bm{x}_n & = &
  \textrm{\code{sum(sum(X .* (X * V)))}} .
\end{eqnarray}
The rest of the code follows closely the update equations derived further above.

\subsubsection{Variational posterior parameters with ARD}

The function \code{vb\_linear\_fit\_ard.m} computes the variational parameters with ARD, and has syntax
\begin{lstlisting}
[w, V, invV, logdetV, an, bn, E_a, L] =
    vb_linear_fit_ard(X, y[, a0, b0, c0, d0])
\end{lstlisting}
where the arguments \code{X} and \code{y}, and optional prior and hyper-prior parameters \code{a0}, \code{b0}, \code{c0}, \code{d0}, have the same structure as for \code{vb\_linear\_fit.m}.
If not given, the prior / hyper-prior parameters default, as for \code{vb\_linear\_fit.m}, to $a_0 = 10^{-2}$, $b_0 = 10^{-4}$, $c_0 = 10^{-2}$, and $d_0 = 10^{-4}$.
The return values \code{w}, \code{V}, \code{invV}, \code{logdetV}, \code{an}, and \code{bn} again correspond to the parameters of the variational posterior $\VarProb^*(\bm{w}, \tau)$, and \code{L} to the variational bound $\varbound(\VarProb)$.
The only difference to \code{vb\_linear\_fit.m} is that \code{E\_a} is a vector with $D$ elements, returning the diagonal elements of $\expect_\alpha(\bm{A})$.

The structure of \code{vb\_linear\_fit.m} is similar to \code{vb\_linear\_fit.m}, updating the parameters of $\VarProb^*(\bm{w}, \tau)$ and $\VarProb^*(\bm{\alpha})$ in lines 67--75 and lines 77--79, respectively, and computing the variational bound, $\varbound(\VarProb)$ in lines 81--84.
This is again repeated until either $\varbound(\VarProb)$ changes by less than $0.001\%$ between two consecutive iterations, or the number of iterations exceeds 500.

\subsubsection{Predictive density parameters}

For a given set of variational posterior parameters, the function \code{vb\_linear\_pred.m} computes the parameters of the predictive density, Eq.~(\ref{eq:vb_linear_pred}), and has syntax
\begin{lstlisting}
[mu, lambda, nu] = vb_linear_pred(X, w, V, an, bn)
\end{lstlisting}
Here \code{X} is the $M \times D$ matrix with $\bm{x}_m^\top$ as its rows.
The other arguments correspond to the variational posterior parameters returned by \code{vb\_linear\_fit.m} or \code{vb\_linear\_fit\_ard.m}.
The function returns the vectors \code{mu} and \code{lambda} with $M$ elements, and the scalar \code{nu}.
These variables specify the mean, precision, and the degrees of freedom of the predictive Student's t distribution for $y_m$ (Eq.~(\ref{eq:vb_linear_pred}), corresponding to $\bm{x}_m$) by the $m$th element of \code{mu} and \code{lambda}, and by \code{nu}, respectively.

\subsection{Examples}

\subsubsection{Estimation of regression coefficients}

Assuming inputs
\begin{lstlisting}
>> X = [ones(100, 1) randn(100, 3)];
>> y = X * [1 2 3 5]' + randn(100, 1);
\end{lstlisting}
The mean regression coefficient estimates are found by
\begin{lstlisting}
>> vb_linear_fit(X, y)
ans =

    0.9269
    2.0220
    2.9915
    4.9798
\end{lstlisting}
Due to the additive noise, these estimates are close to, but do not exactly match the generative coefficients, $\bm{w} = (1, 2, 3, 5)^\top$.

\subsubsection{Higher-dimensional linear regression, and predictive accuracy}

Let us now consider a 100-dimensional case with only 150 observations, that is $D = 100$ and $N = 150$.
We generate the training and testing set by
\begin{lstlisting}
>> D = 100;  N = 150;  N_test = 50;
>> w = randn(D, 1);
>> X = rand(N, D) - 0.5;
>> X_test = rand(N_test, D) - 0.5;
>> y = X * w + randn(N, 1);
>> y_test = X_test * w + randn(N_test, 1);
\end{lstlisting}
such that $\bm{w}$ is drawn from a zero-mean unit variance Gaussian (corresponding to the assumptions of the Bayesian model), and the $\bm{x}_n$'s have elements drawn from a uniform distribution on $[-0.5, 0.5]$.
We train a regression model by both variational Bayesian inference and maximum likelihood by
\begin{lstlisting}
>> [w_VB, V_VB, ~, ~, an_VB, bn_VB] = vb_linear_fit(X, y);
>> y_VB = vb_linear_pred(X, w_VB, V_VB, an_VB, bn_VB);
>> [y_test_VB, lam_VB, nu_VB] = vb_linear_pred(X_test, w_VB, V_VB, an_VB, bn_VB);
>> [w_ML, wint_ML] = regress(y, X);
>> y_ML = X * w_ML;
>> y_test_ML = X_test * w_ML;
\end{lstlisting}
Measuring the mean squared error for both the training and the test set, we find
\begin{lstlisting}
>> fprintf('Training set MSE: ML = %f, VB = %f\n', ...
           mean((y - y_ML).^2), mean((y - y_VB).^2));
Training set MSE: ML = 0.363473, VB = 0.446073
>> fprintf('Test     set MSE: ML = %f, VB = %f\n', ...
           mean((y_test - y_test_ML).^2), mean((y_test - y_test_VB).^2));
Test     set MSE: ML = 3.622844, VB = 3.221452
\end{lstlisting}
Clearly, the maximum likelihood estimate over-fits the training data, as reflected by a small training set error and a large test set error.
Variational Bayesian regression also shows signs of over-fitting, but to a lesser extent than maximum likelihood.

\begin{figure}
  \centering \includegraphics[width=9.5cm]{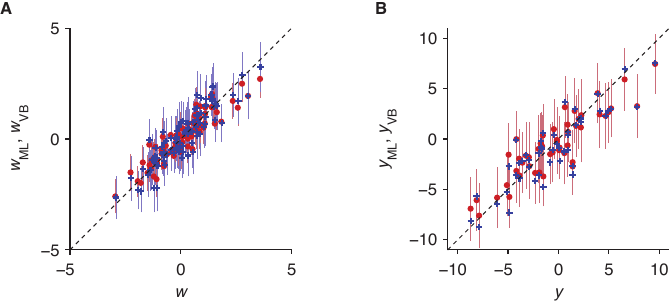}
  \caption{Coefficient estimates and output predictions for 100-dimensional linear regression example.
    A) True coefficients vs.~coefficient estimates (mean $\pm$ 95\% CIs) for variational Bayesian inference (red) and maximum likelihood (blue). The estimates of both inference approaches are horizontally shifted, such that their CIs can be distinguished.
    B) True outputs vs. test set output predictions (mean, $\pm$ 95\% CIs when available) for variational Bayesian inference (red) and maximum likelihood (blue).
    For both coefficients and output predictions, the variational Bayesian estimates are on average closer to the true values.}
  \label{fig:lin_example_highdim}
\end{figure}

We visualize the fit of the regression coefficients by
\begin{lstlisting}
>> figure;  hold on;
>> for i = 1:D
        plot(w(i) * [1 1] - 0.01, w_VB(i) + sqrt(V_VB(i,i)) * 1.96 * [-1 1], ...
             '-', 'LineWidth', 0.25, 'Color', [0.8 0.5 0.5]);
        plot(w(i) * [1 1] + 0.01, wint_ML(i,:), ..., 
             '-', 'LineWidth', 0.25, 'Color', [0.5 0.5 0.8]);
   end
>> plot(w - 0.01, w_VB, 'o', 'MarkerSize', 3, ...
        'MarkerFaceColor', [0.8 0 0], 'MarkerEdgeColor', 'none');
>> plot(w + 0.01, w_ML, '+', 'MarkerSize', 3, ...
        'MarkerFaceColor', 'none', 'MarkerEdgeColor', [0 0 0.8], 'LineWidth', 1);
>> xymin = min([min(xlim) min(ylim)]);  xymax = max([max(xlim) max(ylim)]);
>> plot([xymin xymax], [xymin xymax], 'k--', 'LineWidth', 0.5);
>> set(gca, 'Box','off', 'PlotBoxAspectRatio', [1 1 1], ...
       'TickDir', 'out', 'TickLength', [1 1]*0.02);
>> xlabel('w');  ylabel('w_{ML}, w_{VB}');
\end{lstlisting}
which results in Fig.~\ref{fig:lin_example_highdim}A.
As can be seen, the better fit of variational Bayesian inference is also reflected in a better estimate of the regression coefficients.
Visualizing the test set predictions in the same way results in Fig.~\ref{fig:lin_example_highdim}B.
As for the regression coefficients, variational Bayesian inference can be seen to provide better predictions than maximum likelihood.

The code for this example is available in \code{vb\_linear\_example\_highdim.m}.

\subsubsection{High-dimensional regression with uninformative input dimensions}

To demonstrate the effect of Automated Relevance Determination, consider a high-dimensional input space in which most of the input dimensions are informative (that is, for which the generative regression coefficients are zero).
Specifically, we assume 1000 dimensions, of which only 100 are informative.
We generate training and test data by
\begin{lstlisting}
>> D = 1000;  D_eff = 100;  N = 500;  N_test = 50;
>> w = [randn(D_eff, 1); zeros(D - D_eff, 1)];
>> X = rand(N, D) - 0.5;
>> X_test = rand(N_test, D) - 0.5;
>> y = X * w + randn(N, 1);
>> y_test = X_test * w + randn(N_test, 1);
\end{lstlisting}
Thus, only the first $D_{\textrm{eff}} = 100$ elements of the 1000-dimensional $\bm{w}$ are non-zero.

\begin{figure}
  \centering  \includegraphics[width=9.5cm]{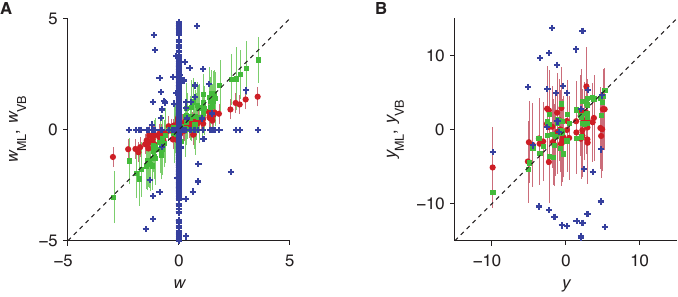}
  \caption{Regression coefficients and predictions for variational Bayesian inference without and with ARD, and maximum likelihood estimation, for 1000-dimensional regression problem with 100 informative dimensions.
    A) True coefficients vs.~coefficient estimates (mean, $\pm$ 95\% CIs where available), computed by maximum likelihood (blue, most outside of plotted range), and by variational Bayesian inference without (red) and with ARD (green).
    While maximum likelihood fails to correctly estimate these coefficients, variational Bayesian inference without ARD applies overly strong shrinkage to informative dimensions, which causes a bias towards small coefficient estimates.
    Only with ARD is it able to modulate the amount of shrinkage applied to different dimensions by their informativeness.
    B) Test set predictions (mean, $\pm$ 95\% CIs where available) vs.~true values, as estimated by maximum likelihood (blue, most outside of plotted range), and by variational Bayesian inference without (red) and with ARD (green).}
  \label{fig:lin_example_sparse}
\end{figure}

We find the coefficients by variational Bayesian inference (without and with ARD) and maximum likelihood by
\begin{lstlisting}
>> [w_VB, V_VB, ~, ~, an_VB, bn_VB] = vb_linear_fit(X, y);
>> y_VB = vb_linear_pred(X, w_VB, V_VB, an_VB, bn_VB);
>> [y_test_VB, lam_VB, nu_VB] = ...
       vb_linear_pred(X_test, w_VB, V_VB, an_VB, bn_VB);
>> [w_VB2, V_VB2, ~, ~, an_VB2, bn_VB2] = vb_linear_fit_ard(X, y);
>> y_VB2 = vb_linear_pred(X, w_VB2, V_VB2, an_VB2, bn_VB2);
>> [y_test_VB2, lam_VB2, nu_VB2] = ...
       vb_linear_pred(X_test, w_VB2, V_VB2, an_VB2, bn_VB2);
>> [w_ML, wint_ML] = regress(y, X);
Warning: X is rank deficient to within machine precision. 
> In regress at 84
> In vb_linear_example_sparse at 37 
>> y_ML = X * w_ML;
>> y_test_ML = X_test * w_ML;
\end{lstlisting}
\proglang{MATLAB}'s \code{regress} function correctly identifies the rank deficiency of $\bm{X}$.
Due to the shrinkage prior on $\bm{w}$, this rank deficiency is not a problem for Bayesian inference.
The resulting mean squared prediction errors are found by
\begin{lstlisting}
>> fprintf('Training set MSE: ML = %f, VB = %f, VB w/ ARD = %f\n', ...
           mean((y - y_ML).^2), mean((y - y_VB).^2), mean((y - y_VB2).^2));
Training set MSE: ML = 0.000000, VB = 0.270214, VB w/ ARD = 0.000000
>> fprintf('Test     set MSE: ML = %f, VB = %f, VB w/ ARD = %f\n', ...
           mean((y_test - y_test_ML).^2), mean((y_test - y_test_VB).^2), ...
           mean((y_test - y_test_VB2).^2));
Test     set MSE: ML = 277.108556, VB = 7.164384, VB w/ ARD = 3.230588
\end{lstlisting}
As in the previous example, the maximum likelihood estimator shows clear signs of over-fitting, as reflected in the large test set error.
Variational Bayesian inference does not suffer from over-fitting to the same extent, as is illustrated by the significantly smaller test set error.
When used with ARD, this training set error shrinks further, which indicates that ARD is better able to identify and ignore uninformative input dimensions.

A look at the regression coefficients in Fig.~\ref{fig:lin_example_sparse}A (plotted as in the previous example) confirms this property.
It illustrates that maximum likelihood was unable to detect the relevant dimensions, whereas variational Bayesian inference without ARD applied overly strong shrinkage to all dimensions.
ARD, in contrast, determined the amount of shrinkage for each input dimension separately, and this way was able selectively suppress a subset of these.
This is also reflected in the model predictions in Fig.~\ref{fig:lin_example_sparse}B, which inference without ARD underestimates due to overly strong shrinkage of its regression coefficient estimates.
With ARD, in contrast, the amount of bias due to shrinkage is reduced.

The code for this example is available in \code{vb\_linear\_example\_sparse.m}.

\subsubsection{Model selection by maximizing variational bound}

An appealing property of variational Bayesian inference is that the variational bound $\varbound(\VarProb)$ lower-bounds the log-model evidence, $\ln \Prob(\data)$, and can thus be used for model selection.
Here, this feature is demonstrated on the basis of finding the order of the polynomial that has generated the observations.
First, let us generate the data by
\begin{lstlisting}
>> D = 3;  N = 10;  D_ML = 6;  Ds = 1:10;
>> x_range = [-5 5];
>> w = randn(D, 1);
>> x = x_range(1) + (x_range(2) - x_range(1)) * rand(N, 1);
>> x_test = linspace(x_range(1), x_range(2), 100)';
>> gen_X = @(x, d) bsxfun(@power, x, 0:(d-1));
>> X = gen_X(x, D);
>> y = X * w + randn(N, 1);
>> y_test = gen_X(x_test, D) * w;
\end{lstlisting}
In the above, $D-1$ determines the order of the generative polynomial (which in this case has 2nd order), $D_{\textrm{ML}}$ specifies the order assumed by the maximum likelihood estimate, and $Ds$ is the range of orders tested by model selection.
We only generate $N = 10$ training data points to make identifying the correct polynomial order difficult.

\begin{figure}
  \centering  \includegraphics[width=9.5cm]{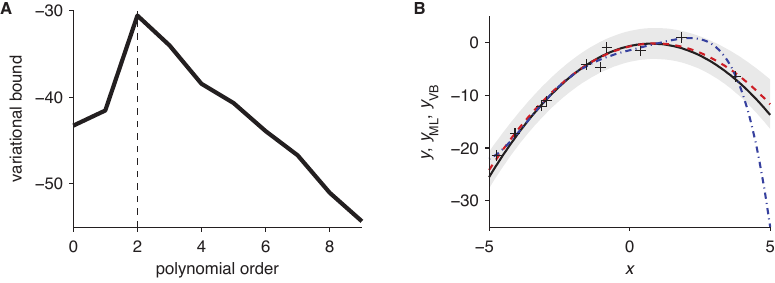}
  \caption{Variational bound and output prediction for identifying the order of the data-generating polynomial by Bayesian model selection.
    A) The variational bound, $\varbound(\VarProb)$, for different generative models, as indexed by the order of the assumed data-generating polynomial.
    This bound lower-bounds the log-model evidence, $\ln \Prob(\data)$, and shows a clear peak at order 2.
    The dashed line indicates the order of the true data-generating polynomial.
    B) Predicted outputs over inputs for the maximum likelihood estimate (blue) and variational Bayesian inference (dashed red).
    The black curve shows the noise-free data-generating polynomial, and the black crosses are the 10 data points based upon which the regression is performed.
    The shaded area indicates the 95\% CIs for the output prediction of the Bayesian model.}
  \label{fig:lin_example_modelsel}
\end{figure}

Model selection is performed by computing the training data variational bound for a set of orders, to find the order that minimizes this bound,
\begin{lstlisting}
Ls = NaN(1, length(Ds));
>> for i = 1:length(Ds)
       [~, ~, ~, ~, ~, ~, ~, Ls(i)] = vb_linear_fit(gen_X(x, Ds(i)), y);
   end
>> [~, i] = max(Ls);
>> D_best = Ds(i)
D_best =

     3
\end{lstlisting}
As can be seen, model selection correctly identified the order of the generative polynomial.
Plotting the variational bound for all tested polynomial orders (Fig.~\ref{fig:lin_example_modelsel}A) shows that this selection was unambiguous.

We test model prediction on the previously generated training set for variational Bayesian inference with the inferred polynomial order, and by maximum likelihood with $D_{ML}$, by
\begin{lstlisting}
>> [w_VB, V_VB, ~, ~, an_VB, bn_VB] = vb_linear_fit(gen_X(x, D_best), y);
>> [y_VB, lam_VB, nu_VB] = ...
       vb_linear_pred(gen_X(x_test, D_best), w_VB, V_VB, an_VB, bn_VB);
>> w_ML = regress(y, gen_X(x, D_ML));
>> y_ML = gen_X(x_test, D_ML) * w_ML;
>> fprintf('Test set MSE, ML = %f, VB = %f\n', ...
           mean((y_test - y_ML).^2), mean((y_test - y_VB).^2));
Test set MSE, ML = 16.601449, VB = 0.603299
\end{lstlisting}
The wrong model ($D_{\textrm{ML}} = 6$ rather than 3) caused the maximum likelihood estimate to perform badly on the test set, as illustrated by its large mean squared error.
Plotting the prediction of both model reveals that the data under-constraints the maximum likelihood, such that its output predictions deviate noticeably from the true outputs for larger inputs (Fig.~\ref{fig:lin_example_modelsel}B).
For the variational Bayesian model, in contrast, the true, generative polynomial remains within the model prediction's 95\% credible intervals.

The code for this example is available in \code{vb\_linear\_example\_modelsel.m}.

\section{Logistic regression}

This section describes how to perform variational Bayesian inference for
logistic regression with a hyper-prior on $\bm{w}$.
The basic generative model is the same as that used in \cite{bishop2006}.
In addition to this, we here provide a variant that performs automatic relevance
determination.

\subsection{The model}

The data $y$ is, dependent on the $D$-dimensional input $\bm{x}$,
assumed to be of either class $y = -1$ or $y = 1$. The log-likelihood
ratio $\ln (\Prob(y = 1 | \bm{x}, \bm{w}) / \Prob(y = -1 | \bm{x}, \bm{w}))$
is assumed to be linear in $\bm{x}$, such that the conditional
likelihood for $y = 1$ is given by the sigmoid
\begin{equation}
  \Prob(y = 1| \bm{x}, \bm{w}) = \frac{1}{1 + \exp(- \bm{w}^\top \bm{x})} =
  \sigma(\bm{w}^\top \bm{x}) .
  \label{eq:logit_model}
\end{equation}
Equally, $\Prob(y = -1 | \bm{x}, \bm{w}) = 1 - \Prob(y = 1 | \bm{x}, \bm{w}) =
1 / (1 + \exp( \bm{w}^\top \bm{x} ))$, such that
\begin{equation}
  \Prob(y | \bm{x}, \bm{w}) = \sigma(y \bm{w}^\top \bm{x}).
\end{equation}

Given some data $\data = \{ \bm{X}, \bm{Y} \}$, where $\bm{X} = \{
\bm{x}_1, \dots, \bm{x}_N \}$ and $\bm{Y} = \{ y_1, \dots, y_N \}$ are
$N$ input/output pairs, the aim is to find the posterior $\Prob(\bm{w} |
\data)$, given some prior $\Prob(\bm{w})$. Unfortunately, the sigmoid data
likelihood does not admit a conjugate-exponential prior. Therefore,
approximations need to be applied to find an analytic expression for
the posterior.

The approximation that will be used is quadratic in $\bm{w}$ in the
exponential, such that the conjugate Gaussian prior
\begin{equation}
  \Prob(\bm{w} | \alpha) = \normaldist(\bm{w} | \bm{0}, \alpha^{-1}
  \bm{I}) = \left( \frac{\alpha}{2 \pi} \right)^{D/2} \exp \left( -
    \frac{\alpha}{2} \bm{w}^\top \bm{w} \right)
\end{equation}
can be used. This prior is parametrized by the hyper-parameter
$\bm{\alpha}$ that is modeled by a conjugate Gamma distribution
\begin{equation}
  \Prob(\alpha) = \gammadist(\alpha | a_0, b_0) =
  \frac{1}{\Gamma(a_0)} b_0^{a_0} \alpha^{a_0 - 1} \exp(-b_0 \alpha) .
  \label{eq:logit_hyperparam}
\end{equation}
This hyper-prior implies the a-prior variance, $\alpha^{-1}$ of the zero-mean $\bm{w}$ to be inverse-gamma with moments $\expect \left( \alpha^{-1} \right) = b_0 / (a_0 - 1)$ for $a_0 > 1$, and $\var \left( \alpha^{-1} \right) = b_0^2 / \left( (a_0 - 1)^2 (a_0 - 2) \right)$ for $a_0 > 2$.

\subsection{Variational Bayesian inference}

Variational Bayesian inference is based on maximizing a lower bound
on the marginal data log-likelihood
\begin{equation}
  \ln \Prob(\bm{Y} | \bm{X}) = \ln \iint \Prob(\bm{Y} | \bm{X}, \bm{w})
  \Prob(\bm{w} | \alpha) \Prob(\alpha) \ud \bm{w} \ud \alpha .
\end{equation}
This lower bound is given by
\begin{equation}
  \ln \Prob(\bm{Y} | \bm{X}) \ge \varbound(\VarProb) = \iint \VarProb(\bm{w}, \alpha)
  \ln \frac{\Prob(\bm{Y} | \bm{X}, \bm{w}) p (\bm{w} | \alpha)
    \Prob(\alpha)}{\VarProb(\bm{w}, \alpha)} \ud \bm{w} \ud \alpha ,
\end{equation}
where the variational distribution $\VarProb(\bm{w}, \alpha)$, approximating
the posterior $\Prob(\bm{w}, \alpha | \data)$, is assumed to factor
into $\VarProb(\bm{w}, \alpha) = \VarProb(\bm{w}) \VarProb(\alpha)$. This approximation
leads to analytic posterior expressions if the model structure is
conjugate-exponential.

The data likelihood
\begin{equation}
  \Prob(\bm{Y} | \bm{X}, \bm{w}) = \prod_n \Prob(y_n | \bm{x}_n, \bm{w})
  = \prod_n \sigma(y_n \bm{w}^\top \bm{x}_n)
\end{equation}
does not admit a conjugate prior in the exponential family and will be
approximated by the use of
\begin{equation}
  \sigma(z) \ge \sigma(\xi) \exp \left( (z - \xi) / 2 -
    \lambda(\xi)(z^2 - \xi^2) \right) , \quad
  \lambda(\xi) = \frac{1}{2 \xi} \left( \sigma(\xi) - \frac{1}{2} \right),
\end{equation}
which is a tight lower bound on the sigmoid, with one additional parameter $\xi$
per datum \citep{jaakkola2000}.
Applying this bound, the data log-likelihood is lower-bounded by
\begin{eqnarray}
  \ln \Prob(\bm{Y} | \bm{X}, \bm{w}) &\ge& \ln h(\bm{w}, \bm{\xi}) \nonumber \\
   &=& \bm{w}^\top \sum_n \frac{y_n}{2} \bm{x}_n - \bm{w}^\top \left( \sum_n
     \lambda(\xi_n) \bm{x}_n \bm{x}_n^\top \right) \bm{w} \nonumber \\
   && \quad + \sum_n \left( \ln \sigma(\xi_n) - \frac{\xi_n}{2} +
     \lambda(\xi_n) \xi_n^2 \right) ,
\end{eqnarray}
with one local variation parameter $\xi_n$ per datum. This results in
the new variational bound
\begin{equation}
  \tilde{\varbound}(\VarProb, \bm{\xi}) = \iint \VarProb(\bm{w}, \alpha) \ln
  \frac{h(\bm{w}, \bm{\xi}) \Prob(\bm{w} | \alpha) \Prob(\alpha)}{\VarProb(\bm{w},
    \alpha)} \ud \bm{w} \ud \alpha ,
\end{equation}
which is a lower bound on the original variational bound, that is
$\tilde{\varbound}(\VarProb, \bm{\xi}) \le \varbound(\VarProb)$.

The variational posteriors are evaluated by standard variational
methods for factorized distributions. The variational posterior for
$\bm{w}$ is given by
\begin{eqnarray}
  \ln \VarProb^*(\bm{w}) &=& \ln h(\bm{w}, \bm{\xi}) + \expect_\alpha( \ln
  \Prob(\bm{w} | \alpha)) + \const\nonumber \\
  &=& \bm{w}^\top \sum_n \frac{y_n}{2} \bm{x}_n - \frac{1}{2} \bm{w}^\top
  \left( \expect_\alpha(\alpha) \bm{I} + 2 \sum_n \lambda(\xi_n)
    \bm{x}_n \bm{x}_n^\top \right) \bm{w} + \const \nonumber  \\
  &=& \ln \normaldist(\bm{w} | \bm{w}_N, \bm{V}_N) ,
  \label{eq:logit_w_post}
\end{eqnarray}
with parameters
\begin{eqnarray}
  \bm{V}_N^{-1} &=& \expect_\alpha(\alpha) \bm{I} + 2 \sum_n
  \lambda(\xi_n) \bm{x}_n \bm{x}_n^\top , \\
  \bm{w}_N &=& \bm{V}_N \sum_n \frac{y_n}{2} \bm{x}_n .
\end{eqnarray}
The variational posterior for $\alpha$ results in
\begin{eqnarray}
  \ln \VarProb^*(\alpha) &=& \expect_{\bm{w}}(\ln \Prob(\bm{w} | \alpha)) + \ln
  \Prob(\alpha) + \const \nonumber \\
  &=& \left( a_0 - 1 + \frac{D}{2} \right) \ln \alpha - \left( b_0 +
  \frac{1}{2} \expect_{\bm{w}} (\bm{w}^\top \bm{w}) \right) \alpha + \const
  \nonumber \\
  &=& \ln \gammadist(\alpha | a_N, b_N) ,
\end{eqnarray}
with
\begin{eqnarray}
  a_N &=& a_0 + \frac{D}{2} , \\
  b_N &=& b_0 + \frac{1}{2} \expect_{\bm{w}}(\bm{w}^\top \bm{w}) .
\end{eqnarray}
The expectations are evaluated with respect to the variational
posteriors and result in
\begin{eqnarray}
  \expect_\alpha(\alpha) &=& \frac{a_N}{b_N} , \\
  \expect_{\bm{w}}(\bm{w}^\top \bm{w}) &=& \bm{w}_N^\top \bm{w}_N +
  \trace(\bm{V}_N) .
\end{eqnarray}

The variational bound itself is given by
\begin{eqnarray}
  \tilde{\varbound}(\VarProb, \bm{\xi}) &=& \expect_{\bm{w}}(\ln h(\bm{w},
  \bm{\xi})) + \expect_{\bm{w}, \alpha}(\ln \Prob(\bm{w} | \alpha)) +
  \expect_\alpha (\ln \Prob(\alpha)) \nonumber \\
  && \quad - \expect_{\bm{w}}(\ln \VarProb(\bm{w})) - \expect_\alpha(\ln
  \VarProb(\alpha)) ,\\
  \expect_{\bm{w}}(\ln h(\bm{w}, \bm{\xi})) &=& \frac{1}{2} \bm{w}_N^\top
  \bm{V}_N^{-1} \bm{w}_N - \frac{D}{2} + \frac{1}{2} \frac{a_N}{b_N}
  \left( \bm{w}_N^\top \bm{w}_N + \trace(\bm{V}_N) \right) \nonumber \\
  && \quad + \sum_n \left( \ln \sigma(\xi_n) - \frac{\xi_n}{2} +
    \lambda(\xi_n) \xi_n^2 \right) , \\
  \expect_{\bm{w}, \alpha}(\ln \Prob(\bm{w} | \alpha)) &=& -
  \frac{D}{2} \ln 2 \pi + \frac{D}{2} ( \psi(a_N) - \ln b_N ) \nonumber \\
  && \quad - \frac{1}{2} \frac{a_N}{b_N} \left(\bm{w}_N^\top \bm{w}_N +
    \trace(\bm{V}_N) \right) , \\
  \expect_\alpha(\ln \Prob(\alpha)) &=& - \ln \Gamma(a_0) + a_0 \ln b_0 +
  (a_0 - 1) ( \psi(a_N) - \ln b_N) - b_0 \frac{a_N}{b_N} , \\
  \expect_{\bm{w}}(\ln \VarProb(\bm{w})) &=& - \frac{1}{2} \ln | \bm{V}_N | -
  \frac{D}{2} (1 + \ln 2 \pi), \\
  \expect_\alpha(\ln \VarProb(\alpha)) &=& - \ln \Gamma(a_N) + (a_N - 1)
  \psi(a_N) + \ln b_N - a_N ,
\end{eqnarray}
where $\psi(\cdot)$ is the digamma function. In combination, this gives
\begin{eqnarray}
  \tilde{\varbound}(\VarProb, \bm{\xi}) &=& \frac{1}{2} \bm{w}_N^\top
  \bm{V}_N^{-1} \bm{w}_N + \frac{1}{2} \ln | \bm{V}_N | + \sum_n
  \left( \ln \sigma(\xi_n) - \frac{\xi_n}{2} + \lambda(\xi_n) \xi_n^2
  \right) \nonumber \\
  && \quad - \ln \Gamma(a_0) + a_0 \ln b_0 - b_0 \frac{a_N}{b_N} -
  a_N \ln b_N + \ln \Gamma(a_N) + a_N .
\end{eqnarray}

This bound is to be maximized in order to find the variational
posteriors for $\bm{w}$ and $\alpha$. The expressions that maximize
this bound with respect to $\VarProb(\bm{w})$ and $\VarProb(\bm{\alpha})$, while
keeping all other parameters fixed, are given by $\VarProb^*(\bm{w})$ and
$\VarProb^*(\alpha)$ respectively. To find the local variational parameters
$\xi_n$ that maximize $\tilde{\varbound}(\VarProb, \bm{\xi})$, its derivative
with respect to $\xi_n$ is set to zero (see \cite{bishop2006}), resulting
in
\begin{equation}
  (\xi_n^\textrm{new})^2 = \bm{x}_n^\top \left( \bm{V}_N + \bm{w}_N
    \bm{w}_N^\top \right) \bm{x}_n .
\end{equation}
The variational bound is maximized by iterating over the update
equations for $\bm{w}_N$, $\bm{V}_N$, $a_N$, $b_N$ and $\bm{\xi}$,
until $\tilde{\varbound}(\VarProb, \bm{\xi})$ reaches a plateau. A lower
bound on the marginal data log-likelihood $\ln \Prob(\data)$ is given by the
variational bound itself, as $\ln \Prob(\data) \ge \varbound(\VarProb) \ge
\tilde{\varbound}(\VarProb, \bm{\xi}).$

\subsection{Predictive density}

In order to get the predictive density, the posterior $\Prob(\bm{w} |
\data)$ is approximated by the variational posterior $\VarProb(\bm{w})$, and
the sigmoid is lower-bounded by above bound, such that
\begin{eqnarray}
  \Prob(y = 1 | \bm{x}, \data) &=& \int \Prob(y = 1 | \bm{x}, \bm{w}) \Prob(\bm{w}
  | \data) \ud \bm{w} \nonumber \\
  &\approx& \int \Prob(y = 1 | \bm{x}, \bm{w}) \VarProb(\bm{w}) \ud \bm{w} , \\
  &\ge& \int \sigma(\xi) \exp \left( \frac{\bm{w}^\top \bm{x} -
      \xi}{2} - \lambda(\xi) \bm{w}^\top \bm{x} \bm{x}^\top \bm{w} +
    \lambda(\xi) \xi^2 \right) \VarProb(\bm{w}) \ud \bm{w} . \nonumber
\end{eqnarray}
The integral is solved by noting that the lower bound is exponentially
quadratic in $\bm{w}$, such that the Gaussian can be completed, to give
\begin{equation}
  \ln \Prob(y = 1 | \bm{x}, \data) \approx \frac{1}{2} \ln
  \frac{|\tilde{\bm{V}}|}{|\bm{V}_N|} - \frac{1}{2} \bm{w}_N^\top
  \bm{V}_N^{-1} \bm{w}_N + \frac{1}{2} \tilde{\bm{w}}^\top
  \tilde{\bm{V}}^{-1} \tilde{\bm{w}} + \ln \sigma(\xi) - \frac{\xi}{2}
  +\lambda(\xi) \xi^2  ,
\end{equation}
with
\begin{eqnarray}
  \tilde{\bm{V}}^{-1} &=& \bm{V}_N^{-1} + 2 \lambda(\xi) \bm{x}
  \bm{x}^\top, \\
  \tilde{\bm{w}} &=& \tilde{\bm{V}} \left( \bm{V}_N^{-1} \bm{w}_N +
    \frac{\bm{x}}{2} \right) .
\end{eqnarray}
The bound parameter $\xi$ that maximizes $\ln \Prob(y = 1 | \bm{x},
\data)$ is given by
\begin{equation}
  (\xi^{\textrm{new}})^2 = \bm{x}^\top \left( \tilde{\bm{V}} +
    \tilde{\bm{w}} \tilde{\bm{w}}^\top \right) \bm{x} .  
\end{equation}
Thus, the predictive density is found by iterating over the updates
for $\tilde{\bm{w}}$, $\tilde{\bm{V}}$ and $\xi$ until $\ln \Prob(y = 1 |
\bm{x}, \data)$ reaches a plateau. The hyper-prior $\Prob(\alpha)$ does
not need to be considered as it does not appear in the variational
posterior $\VarProb(\bm{w})$.

\subsection{Using automatic relevance determination}

To use automatic relevance determination (ARD), each element of the
prior of $\bm{w}$ is assigned a separate prior,
\begin{equation}
  \Prob(\bm{w} | \bm{\alpha}) = \normaldist(\bm{w} | \bm{0}, \bm{A}^{-1})
  = \frac{|\bm{A}|^{1/2}}{\sqrt{2 \pi}^D} \exp \left( - \frac{1}{2}
    \bm{w}^\top \bm{A} \bm{w} \right) ,
\end{equation}
where $\bm{A}$ is the diagonal matrix with the vector $\bm{\alpha} =
(\alpha_1, \dots, \alpha_D)^\top$ along its diagonal. The conjugate
hyper-prior $\Prob(\bm{\alpha})$ is given by
\begin{equation}
  \Prob(\bm{\alpha}) = \prod_i \gammadist(\alpha_i | a_0, b_0) .
\end{equation}
Note that $\alpha_i$ determines the precision (inverse variance) of
the $i$th element of $\bm{w}$. A low precision makes the prior
uninformative, whereas a high precision tells us that the associated
element in $\bm{w}$ is most likely zero and the associated input
element is therefore irrelevant for the prediction of $y$. Thus, such a
prior structure automatically determines the relevance of each element
of the input to predict the class of the output.

Using the same variational Bayes inference as before, the variational
posteriors are given by
\begin{equation}
  \VarProb^*(\bm{w}) = \normaldist(\bm{w} | \bm{w}_N, \bm{V}_N), \quad
  \VarProb^*(\bm{\alpha}) = \prod_i \gammadist(\alpha_i | a_N, b_{Ni}) ,
\end{equation}
with
\begin{eqnarray}
  \bm{V}_N^{-1} &=& \expect_\alpha(\bm{A}) + 2 \sum_n \lambda(\xi_n)
  \bm{x}_n \bm{x}_n^\top , \\
  \bm{w}_N &=& \bm{V}_N \sum_n \frac{y_n}{2} \bm{x}_n, \\
  a_N &=& a_0 + \frac{1}{2} , \\
  b_{Ni} &=& b_0 + \frac{1}{2} \expect_{\bm{w}}(w_i^2) ,
\end{eqnarray}
where $w_i$ is the $i$th element of $\bm{w}$, and $\bm{A}_N =
\expect_\alpha(\bm{A})$ is a diagonal matrix with its $i$th diagonal
element given by $\expect_\alpha(\alpha_i) = a_N /
b_{Ni}$. $\expect_{\bm{w}}(w_i^2)$ evaluates to $\expect_{\bm{w}}(w_i^2) =
\expect_{\bm{w}}(w_i)^2 + \var_{\bm{w}}(w_i) = (\bm{w}_N)_i^2 +
(\bm{V}_N)_{ii}$.

The new expectations to evaluate the variation bound are
\begin{eqnarray}
  \expect_{\bm{w}}(\ln h(\bm{w}, \bm{\xi})) &=& \frac{1}{2} \bm{w}_N^\top
  \bm{V}_N^{-1} \bm{w}_N - \frac{D}{2} + \frac{1}{2} \left(
    \trace(\bm{A}_N \bm{V}_N) + \bm{w}_N^\top \bm{A}_N \bm{w}_N \right)
  \nonumber \\
  && \quad + \sum_n \left( \ln \sigma(\xi_n) - \frac{\xi_n}{2} +
    \lambda(\xi_n) \xi_n^2 \right) , \\
  \expect_{\bm{w}, \bm{\alpha}}(\ln \Prob(\bm{w} | \bm{\alpha})) &=&
  \frac{1}{2} \sum_i \left( \psi(a_N) - \ln b_{Ni} \right) \nonumber \\
  && \quad - \frac{D}{2} \ln 2 \pi - \frac{1}{2} \left(
    \trace(\bm{A}_N \bm{V}_N) + \bm{w}_N^\top \bm{A}_N \bm{w}_N \right) ,
  \\
  \expect_{\bm{\alpha}}(\ln \Prob(\bm{\alpha})) &=& \sum_i \Big( - \ln
    \Gamma(a_0) + a_0 \ln b_0 \nonumber \\
  && \quad \quad + (a_0 - 1) ( \psi(a_N) - \ln b_{Ni}) -
    b_0 \frac{a_N}{b_{Ni}} \Big) , \\
  \expect_{\bm{\alpha}}(\ln \VarProb(\bm{\alpha})) &=& \sum_i \left( - \ln
    \Gamma(a_N) + (a_N - 1) \psi(a_N) + \ln b_{Ni} - a_N \right) ,
\end{eqnarray}
resulting in
\begin{eqnarray}
  \tilde{\varbound}(\VarProb, \bm{\xi}) &=& \frac{1}{2} \bm{w}_N^\top
  \bm{V}_N^{-1} \bm{w}_N + \frac{1}{2} \ln | \bm{V}_N | + \sum_n
  \left( \ln \sigma(\xi_n) - \frac{\xi_n}{2} + \lambda(\xi_n) \xi_n^2
  \right) \\
  && \quad + \sum_i \Big( - \ln \Gamma(a_0) + a_0 \ln b_0 - b_0
    \frac{a_N}{b_{Ni}} - a_N \ln b_{Ni} + \ln \Gamma(a_N) + a_N
  \Big) . \nonumber
\end{eqnarray}
As the variational posterior $\VarProb^*(\bm{w})$ is independent
of the hyper-parameters, the predictive density is evaluated as
before.

\subsection{Implementation}

The scripts that compute the variational posterior parameters without and with ARD are \code{vb\_logit\_fit.m} and \code{vb\_logit\_fit\_ard.m}, respectively.
\code{vb\_logit\_fit\_iter.m} is a slower version of \code{vb\_logit\_fit.m} that features a slightly simplified generative model without the hyper-prior, and iterates over updating each $\xi_n$ separately rather updating them for all $\bm{x}_n$'s simultaneously.
\code{vb\_logit\_pred.m} computes the predictive density parameters for a set of input vectors.
\code{vb\_logit\_pred\_iter.m} does so as well, but again iterates over updating $\xi_n$ rather than updating them all at the same time. 

\subsubsection{Variational posterior parameters without ARD, iterative implementation}

The function \code{vb\_logit\_fit\_iter.m} deviates from the generative model given by Eqs.~(\ref{eq:logit_model})-(\ref{eq:logit_hyperparam}) as it does not use a hyper-prior on $\alpha$.
Instead, it uses the conjugate Gaussian prior
\begin{equation}
  \Prob(\bm{w}) = \normaldist \left( \bm{w} | \bm{0}, D^{-1} \bm{I} \right) ,
\end{equation}
where $D$ is the input dimensionality.
This prior was chosen to increase shrinkage with the number of input dimensions.
The function is called by
\begin{lstlisting}
[w, V, invV, logdetV] = vb_logit_fit_iter(X, y)
\end{lstlisting}
where \code{X} is the $N \times D$ input matrix with $\bm{x}_n^\top$ as its $n$th row, and \code{y} is the output column vector with $N$ elements that are either $-1$ or $1$.
The returned \code{w}, \code{V} specify the parameters $\bm{w}_N$, $\bm{V}_N$ of the variational posterior Eq.~(\ref{eq:logit_w_post}).
The function additionally returns $\textrm{\code{invV}} = \bm{V}_N^{-1}$ and $\textrm{\code{logdetV}} = \ln | \bm{V}_n |$, such that, if required, these values do not need to be re-computed.

The function computes these parameters incrementally by adding the observations $\bm{x}_n, y_n$ one by one, while optimizing $\xi_n$ for each of these observations separately. 
Let $\bm{V}_j$ and $\bm{w}_j$ denote the parameters of $\VarProb^*(\bm{w})$ after $j$ observations have been made.
Starting with $\bm{w}_0 = \bm{0}$, $\bm{V}_0 = D^{-1} \bm{I}$, $\bm{V}_0^{-1} = D \bm{I}$, and $\ln | \bm{V}_0^{-1} | = -D \ln D$ according to the prior, $\bm{V}_j$ follows the incremental update
\begin{equation}
  \bm{V}_j^{-1} = \bm{V}_{j-1}^{-1} + 2 \lambda(\xi_j) \bm{x}_j
  \bm{x}_j^\top .
\end{equation}
The incremental update of $\bm{w}_j$ is slightly more complex, but from observing that
\begin{equation}
  \bm{V}_j^{-1} \bm{w}_j = \sum_n^j \frac{y_n}{2} \bm{x}_n =
  \frac{y_j}{2} \bm{x}_j + \sum_n^{j - 1} \frac{y_n}{2} \bm{x}_n =
  \bm{V}_{j-1}^{-1} \bm{w}_{j-1} + \frac{y_j}{2} \bm{x}_j ,
\end{equation}
it is easy to see that
\begin{equation}
  \bm{w}_j = \bm{V}_j \left( \bm{V}_{j-1}^{-1} \bm{w}_{j-1} +
    \frac{y_j}{2} \bm{x}_j \right) .
\end{equation}
The script avoids taking the inverse of $\bm{V}$ by updating $\bm{V}^{-1}$ and $\bm{V}$ in parallel, where the latter is based on an application of the Sherman-Morrison formula on the $\bm{V}^{-1}$ update, resulting in
\begin{equation}
  \bm{V}_j = \left( \bm{V}_{j-1}^{-1} + 2 \lambda(\xi_j) \bm{x}_j
    \bm{x}_j^\top \right)^{-1} = \bm{V}_{j-1} - \frac{2 \lambda(\xi_j)
    \bm{V}_{j-1} \bm{x}_j \bm{x}_j^\top \bm{V}_{j-1}}{1 + 2
    \lambda(\xi_j) \bm{x}_j^\top \bm{V}_{j-1} \bm{x}_j} .
\end{equation}
$\ln | \bm{V}_j |$ can be updated in a similar way, based on the Matrix determinant lemma,
\begin{equation}
  | \bm{V}_j^{-1} | = | \bm{V}_{j-1}^{-1} + 2 \lambda(\xi_j) \bm{x}_j
  \bm{x}_j^\top | = | \bm{V}_{j-1}^{-1} | \left( 1 + 2 \lambda(\xi_j)
  \bm{x}_j^\top \bm{V}_{j-1} \bm{x}_j \right) ,
\end{equation}
such that, using $\ln | \bm{V}_j | = - \ln | \bm{V}_j^{-1} |$,
\begin{equation}
  \ln | \bm{V}_j | = \ln | \bm{V}_{j-1} | - \ln \left( 1 + 2
    \lambda(\xi_j) \bm{x}_j^\top \bm{V}_{j-1} \bm{x}_j \right) .
\end{equation}
The function initializes $\bm{V}_0$ and $\bm{w}_0$ in lines 41--44, and then updates its values by iterating over the $\bm{x}_j$'s for $j = 1, \dots, N$, starting in line 48.
For each $j$, it first updates $\bm{V}$ and $\bm{w}$ under the assumption that $\xi_n = 0$ and $\lambda(\xi_n) = 1 /8$, leading to a simplified initial step,
\begin{eqnarray}
  \bm{V}_j^{-1}(\xi_j) & =_{\xi_j = 0} & \bm{V}_{j-1}^{-1} + \frac{1}{4}
  \bm{x}_j \bm{x}_j^\top , \\
  \bm{V}_j(\xi_j) & =_{\xi_j = 0} & \bm{V}_{j-1}-\frac{\bm{V}_{j-1}
    \bm{x}_j \bm{x}_j^\top \bm{V}_{j-1}}{4 + \bm{x}_j^\top \bm{V}_{j-1}
    \bm{x}_j} , \\
  \ln | \bm{V}_j(\xi_j) | & =_{\xi_j = 0} & \ln | \bm{V}_{j-1} | -
  \ln \left(1 + \frac{1}{4} \bm{x}_j^\top \bm{V}_{j-1} \bm{x}_j \right) .
\end{eqnarray}
Then, in lines 66--90, it alternates between updating $\xi_j$, and $\bm{V}_j(\xi_j)$ and $\bm{w}_j(\xi_j)$ while monitoring how these updates change the variational bound $\varbound(\xi_j)$.
The updates are performed until the variational bound changes less than 0.001\% between two consecutive updates of all parameters, or the number of iterations exceeds 500.
The variational bound itself is, without the hyper-prior, given by
\begin{equation}
  \varbound_j(\xi_j) = \frac{1}{2} \bm{w}_j^\top(\xi_j)
  \bm{V}_j^{-1}(\xi_j) \bm{w}_j(\xi_j) + \frac{1}{2} \ln |
  \bm{V}_j(\xi_j) | + \ln \sigma(\xi_j) - \frac{\xi_j}{2} +
  \lambda(\xi_j) \xi_j^2 .
\end{equation}

\subsubsection{Variational posterior parameters without ARD, batch implementation}

Rather than updating all $\xi_n$ in turn, \code{vb\_logit\_fit.m} estimates the variational posterior parameters by updating all $\xi_n$'s at once.
Furthermore, it differs from \code{vb\_logit\_fit\_iter.m} in that it assumes the full generative mode, Eqs.~(\ref{eq:logit_model})-(\ref{eq:logit_hyperparam}), including the hyper-prior on $\alpha$ with associated parameters $a$ and $b$.
The function is called by
\begin{lstlisting}
[w, V, invV, logdetV, E_a, L] = vb_logit_fit(X, y[, a0, b0])
\end{lstlisting}
where \code{X} and \code{y} specify inputs and outputs as for \code{vb\_logit\_fit\_iter.m}.
The optional \code{a0} and \code{b0} specify the hyper-prior parameters $a_0$ and $b_0$.
If not given, they default to $a_0 = 10^{-2}$ and $b_0 = 10^{-4}$, corresponding to an un-informative hyper-prior (see Sec.~\ref{sec:vblin_implementation}).
The returned \code{w} and \code{V} correspond to the posterior parameters $\bm{w}_N$ and $\bm{V}_N$ of the variational posterior of $\bm{w}$.
\code{E\_a} is $\expect_\alpha(\alpha)$ of the posterior $\alpha$, and \code{L} is the variational bound $\tilde{\varbound}(\VarProb, \bm{\xi})$ at these parameters and the last-used $\bm{\xi}$.
The function additionally returns $\textrm{\code{invV}} = \bm{V}_N^{-1}$ and $\textrm{\code{logdetV}} = \ln | \bm{V}_N |$, such that, if required, these values do not need to be re-computed.

Within the function, all $\xi_n$ are stored in the vector \code{xi} and are updated simultaneously.
The script start in line 54 by assuming $\xi_n = 0$ for all $n$, such that $\lambda(\xi_n) = 1/8$.
Additionally, it pre-computes $\textrm{\code{w\_t}} = \sum_n \bm{x}_n y_n / 2$.
The initial update of $\bm{V}_N(\bm{\xi})$, $\bm{w}_N(\bm{\xi})$, $b_N(\bm{\xi})$, and $\tilde{\varbound}(\VarProb, \bm{\xi})$ in lines 56--62 is computed outside of the loop.
After that, the script iterates in lines 66--95 over first updating $\bm{\xi}$, then $b_N(\bm{\xi})$, followed by $\bm{V}_N(\bm{\xi})$ and $\bm{w}_N(\bm{\xi})$.
The iteration stops if either $\tilde{\varbound}(\VarProb, \bm{\xi})$ does not change more than $0.001\%$ between two consecutive iterations, or the number of iterations exceeds 500.

The script employs a few short-cuts and vectorizations, which will be discussed here.
In particular, the initial $\tilde{\varbound}(\VarProb, \bm{\xi})$ at $\bm{\xi} = \bm{0}$ is simplified by using
\begin{equation}
  \sum_n \left( \ln \sigma(\xi_n) - \frac{\xi_n}{2} + \lambda(\xi_n)
    \xi_n^2 \right) =_{\bm{\xi} = \bm{0}} - N \ln 2 .
\end{equation}
Also, as the scripts computes $\bm{V}_N(\bm{\xi})$ by inverting $\bm{V}_N^{-1}(\bm{\xi})$, it computes $\ln | \bm{V}_N(\bm{\xi}) |$ from $\bm{V}_N^{-1}(\bm{\xi})$ for better stability, using $\ln | \bm{V}_N(\bm{\xi}) | = - \ln | \bm{V}_N^{-1}(\bm{\xi}) |$.
In addition, the following vectorized operations are used:
\begin{eqnarray}
  \sum_n \frac{y_n}{2} \bm{x}_n &=& \textrm{\code{0.5 *
      sum(bsxfun(@times, X, y), 1)'}}, \\
  2 \sum_n \lambda(\xi_n) \bm{x}_n \bm{x}_n^\top &=& \textrm{\code{2
    * X' * bxsfun(@times, X, lam\_xi)}}, \\
  \bm{x}_n^\top \left( \bm{V}_N + \bm{w}_N \bm{w}_N^\top \right) \bm{x}_n
  &=& \left( \textrm{\code{sum(X .* (X * (V + w * w')), 2)}}
  \right)_n . 
\end{eqnarray}
Using a hyper-prior comes at the cost of having to explicitly invert $\bm{V}^{-1}$ at each iteration.
For this reason, \code{vb\_logit\_fit.m} might be numerically less stable than \code{vb\_logit\_fit\_iter.m} for problems will ill-conditioned inputs.

\subsubsection{Variational posterior parameters with ARD}

Estimating the variational posterior parameters with ARD is implemented in the function \code{vb\_logit\_fit\_ard.m}, which is called by
\begin{lstlisting}
[w, V, invV, logdetV, E_a, L] = vb_logit_fit_ard(X, y[, a0, b0])
\end{lstlisting}
The inputs \code{X} and \code{y}, the optional inputs \code{a0} and \code{b0}, and the outputs \code{w}, \code{V}, \code{invV}, \code{logdetV}, and \code{L} have the same meaning as for \code{vb\_logit\_fit.m}.
The only difference is that the returned \code{E\_a} is now a vector that contains the posterior $\expect_{\bm{\alpha}}(\alpha_i)$'s as its elements.

The implementation of \code{vb\_logit\_fit\_ard.m} differs from \code{vb\_logit\_fit.m} only by the variables \code{b\_n} and \code{E\_a}, holding the $b_{Ni}$'s and $\expect_{\bm{\alpha}}(\alpha_i)$'s, now being vectors rather than scalars.
Their update and use is adjusted accordingly, in line with what has been described above.

\subsubsection{Predictive density parameters}

Two scripts are available to compute the predictive probability $\Prob(y = 1 | \bm{x}, \data)$, namely \code{vb\_logit\_pred\_iter.m} and \code{vb\_logit\_pred.m}.
Their only difference is that the latter is a vectorized form of the former.
Both are called by
\begin{lstlisting}
out = vb_logit_pred[_incr](X, w, V, invV)
\end{lstlisting}
where \code{X} is the $M \times D$ matrix with the input vector $\bm{x}_m^\top$ as its $m$th row.
The variational posterior parameters \code{w}, \code{V}, and \code{invV} are those returned by either of the estimation function discussed above.
The returned vector \code{out} with $M$ elements contains the estimated $\Prob(y_m = 1 | \bm{x}_m, \data)$ as its $m$th element.

In terms of implementation, let us first consider \code{vb\_logit\_pred\_iter.m}.
This function iterates over all given $\bm{x}_m$, and optimizes $\bm{\xi}_m$ for each of these separately.
In order to do so, it employs a simplification to $\textrm{\code{logdetV\_xi}} = \ln |\tilde{\bm{V}}| / |\bm{V}_N|$, appearing in $\ln \Prob(y = 1 | \bm{x}, \data)$.
From the expression for $\tilde{\bm{V}}^{-1}$ and the Matrix determinant lemma it can be shown that
\begin{eqnarray}
  \ln | \tilde{\bm{V}} | &=& - \ln | \tilde{\bm{V}}^{-1} | \nonumber \\
  &=& - \ln | \bm{V}_N^{-1} | - \ln \left( 1 + 2 \lambda(\xi) \bm{x}^\top \bm{V}_N
  \bm{x}^\top \right) \nonumber \\
  &=& \ln | \bm{V}_N | - \ln \left(1 + 2 \lambda(\xi) \bm{x}^\top
  \bm{V}_N \bm{x} \right) .
\end{eqnarray}
Thus, $\ln |\tilde{\bm{V}}| / |\bm{V}_N|$ results in
\begin{equation}
  \ln \frac{| \tilde{\bm{V}} |}{| \bm{V}_N |} = - \ln \left( 1 + 2
    \lambda(\xi) \bm{x}^\top \bm{V}_N \bm{x} \right)
\end{equation}
Additionally, the Sherman-Morrison formula can be applied to avoid inverting $\tilde{\bm{V}}^{-1}$ by using
\begin{equation}
  \tilde{\bm{V}} = \bm{V}_N - \frac{2 \lambda(\xi) \bm{V}_N \bm{x}
    \bm{x}^\top \bm{V}_N}{1 + 2 \lambda(\xi) \bm{x}^\top \bm{V}_N \bm{x}}
\end{equation}
instead.

The script initially starts with $\xi_m = 0$ for each $\bm{x}_m$ in lines 59--60, using some initial simplifications based on $\lambda(\xi) = 1/8$, as already previously discussed.
It then iterates over updating $\xi$, $\tilde{\bm{V}}$ and $\tilde{\bm{w}}$ in lines 66--90 until the variational bound either changes less than $0.001\%$ between two consecutive iterations, or the number of iterations exceeds 500.
The variational bound is computed as a simplified version of $\ln \Prob(y = 1| \bm{x}, \data)$, omitting all terms that are independent of $\xi_m$.

The vectorized script \code{vb\_logit\_pred.m} follows exactly the same principles, but optimizes $\bm{\xi}$ for all $\bm{x}$ at the same time, by maximizing a sum of the individual variational bounds.

\subsection{Examples}

\subsubsection{Estimation of coefficients and separating hyperplane}

As a first example, consider 50 observations from a 3-dimensional input space, such that $N = 50$ and $D = 3$. The observations are generated by
\begin{lstlisting}
>> D = 3;  N = 100;  N_test = 1000;  X_scale = 5;
>> w = randn(D, 1);
>> X = [ones(N, 1) (X_scale*(rand(N, 1)-0.5))];
>> X = [X (X_scale*(rand(N, 1)-0.5) - (w(1) + X(:,2) * w(2)) / w(3))];
>> X_test = [ones(N_test, 1) (X_scale*(rand(N_test, 1)-0.5))];
>> X_test = [X_test (X_scale*(rand(N_test, 1)-0.5) - ...
                     (w(1) + X_test(:,2) * w(2)) / w(3))];
>> py = 1 ./ (1 + exp(- X * w));
>> y = 2 * (rand(N, 1) < py) - 1;
>> y_test = 2 * (rand(N_test, 1) < 1 ./ (1 + exp(- X_test * w))) - 1;
\end{lstlisting}
The different $\bm{x}_n$'s are generated such that roughly half of all outputs are 1 and the other half are -1.
This is achieved by setting the first element, $x_{n1}$, of each $\bm{x}_n$ to one, and by drawing the second, $x_{n2} \sim \unifdist(-2.5, 2.5)$ from a uniform distribution over the range $[-2.5, 2.5]$.
The third element is then set to $\unifdist(-2.5, 2.5) - (w_1 + x_{n2} w_2) / w_3$, such that, with 50\% change, $w_1 + x_{n2} w_2 + x_{n3} w_3 > 0$.
The outputs are drawn from $\{-1, 1\}$ according to their probabilities of being 1.

\begin{figure}
  \centering  \includegraphics[width=9.5cm]{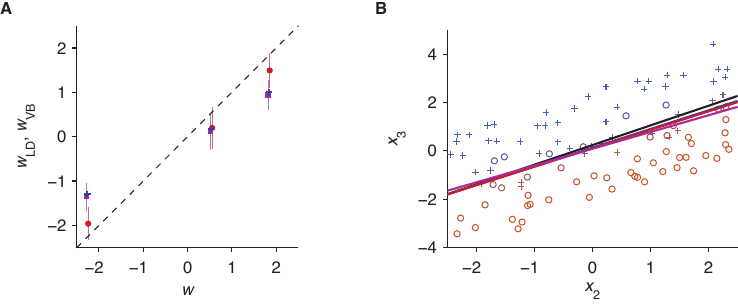}
  \caption{Estimated coefficients and separating hyperplanes for classifiers trained on low-dimensional training set. A) true coefficients vs. their estimates (mean, $\pm$ 95\% CIs where available) for variational Bayesian logistic regression with hyper-prior on $\bm{w}$ (red), variational Bayesian logistic regression without such hyper-prior (purple), and Fisher's linear discriminant (blue). B) Separating hyperplanes (defined by $w_1 + x_2 w_2 + x_3 w_3 = 0$) for the three classifiers (same colors) and generative separating hyperplane (black). The shown observations (circles / crosses) represent the training data, with symbol type indicating class membership, symbol color indicating class probability (blue barely visible, as hidden by red line).}
  \label{fig:logit_example_coeff}
\end{figure}

We train two variational Bayesian logistic regression classifiers, one with and one without a hyper-prior on $\bm{w}$
\begin{lstlisting}
>> [w_VB, V_VB, invV_VB] = vb_logit_fit(X, y);
>> py_VB = vb_logit_pred(X, w_VB, V_VB, invV_VB);
>> py_test_VB = vb_logit_pred(X_test, w_VB, V_VB, invV_VB);
>> [w_VB1, V_VB1, invV_VB1] = vb_logit_fit_iter(X, y);
>> py_VB1 = vb_logit_pred(X, w_VB1, V_VB1, invV_VB1);
>> py_test_VB1 = vb_logit_pred(X_test, w_VB1, V_VB1, invV_VB1);
\end{lstlisting}
In addition, we train Fisher's linear discriminant \citep{bishop2006} by
\begin{lstlisting}
>> y1 = y == 1;
>> w_LD = NaN(D, 1);
>> w_LD(2:end) = (cov(X(y1, 2:end)) + cov(X(~y1, 2:end))) \ ...
                 (mean(X(y1, 2:end))' - mean(X(~y1, 2:end))');
>> w_LD(1) = - 0.5 * (mean(X(y1, 2:end)) + ...
                      mean(X(~y1, 2:end))) * w_LD(2:end);
>> y_LD = 2 * (X * w_LD > 0) - 1;
>> y_test_LD = 2 * (X_test * w_LD > 0) - 1;
\end{lstlisting}
These classifiers feature training and test set errors
\begin{lstlisting}
>> fprintf('training set 0-1 loss: FLD = %f, VB = %f, VBiter = %f\n', ...
           mean(y_LD ~= y), ...
           mean(2 * (py_VB > 0.5) - 1 ~= y), ...
           mean(2 * (py_VB1 > 0.5) - 1 ~= y));
training set 0-1 loss: FLD = 0.140000, VB = 0.140000, VBiter = 0.160000
>> fprintf('test     set 0-1 loss: FLD = %f, VB = %f, VBiter = %f\n', ...
           mean(y_test_LD ~= y_test), ...
           mean(2 * (py_test_VB > 0.5) - 1 ~= y_test), ...
           mean(2 * (py_test_VB1 > 0.5) - 1 ~= y_test));
test     set 0-1 loss: FLD = 0.142000, VB = 0.143000, VBiter = 0.149000
\end{lstlisting}
As can be seen, for this realization of training and test set, all classifiers perform roughly equally well.
Their similar performance is also reflected in the similarity of their coefficient estimates (Fig.~\ref{fig:logit_example_coeff}A) and their class-separating hyperplanes (Fig.~\ref{fig:logit_example_coeff}B).

The code for this example is available in \code{vb\_logit\_example\_coeff.m}.

\subsubsection{High-dimensional logistic regression with uninformative input dimensions}

To illustrate the use of shrinkage priors on all/individual dimensions, let us consider an example in which the majority of input dimensions are uninformative.
We generate the data by
\begin{lstlisting}
>> D = 1000;  D_eff = 100;  N = 2000;  N_test = 10000;
>> w = [randn(D_eff, 1); zeros(D - D_eff, 1)];
>> X = rand(N, D) - 0.5;
>> X_test = rand(N_test, D) - 0.5;
>> py = 1 ./ (1 + exp(- X * w));
>> y = 2 * (rand(N, 1) < py) - 1;
>> py_test = 1 ./ (1 + exp(-X_test * w));
>> y_test = 2 * (rand(N_test, 1) < py_test) - 1;
\end{lstlisting}
In this case, only the first $D_{\textrm{eff}} = 100$ out of the $D = 1000$ elements of each $\bm{x}_n$ are informative about the class label $y_n$.

As before, we train a variational Bayesian logistic regression classifier without ARD, and with and without hyper-prior on $\bm{w}$, and additionally one classifier with ARD, by
\begin{lstlisting}
>> [w_VB, V_VB, invV_VB] = vb_logit_fit(X, y);
>> py_VB = vb_logit_pred(X, w_VB, V_VB, invV_VB);
>> py_test_VB = vb_logit_pred(X_test, w_VB, V_VB, invV_VB);
>> [w_VB1, V_VB1, invV_VB1] = vb_logit_fit_iter(X, y);
>> py_VB1 = vb_logit_pred(X, w_VB1, V_VB1, invV_VB1);
>> py_test_VB1 = vb_logit_pred(X_test, w_VB1, V_VB1, invV_VB1);
>> [w_VB2, V_VB2, invV_VB2] = vb_logit_fit_ard(X, y);
>> py_VB2 = vb_logit_pred(X, w_VB2, V_VB2, invV_VB2);
>> py_test_VB2 = vb_logit_pred(X_test, w_VB2, V_VB2, invV_VB2);
\end{lstlisting}
Furthermore, we train Fisher's linear discriminant by
\begin{lstlisting}
>> y1 = y == 1;
>> w_LD = (cov(X(y1, :)) + cov(X(~y1, :))) \ ...
          (mean(X(y1, :))' - mean(X(~y1, :))');
>> c_LD = 0.5 * (mean(X(y1, :)) + mean(X(~y1, :))) * w_LD;
>> y_LD = 2 * (X * w_LD > c_LD) - 1;
>> y_test_LD = 2 * (X_test * w_LD > c_LD) - 1;
\end{lstlisting}
The above computation of the linear discriminant differs slightly from the previous example, as here, no offset term was included in the input matrix $\bm{X}$.

\begin{figure}
  \centering \includegraphics[width=9.5cm]{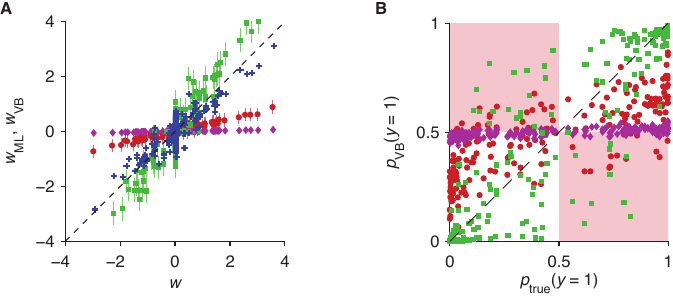}
  \caption{Coefficient estimates and output predictions for high-dimensional classification with few informative dimensions.
    A) true coefficients vs. coefficient estimates (mean, $\pm$ 95\% CIs where available) for Fisher's linear discriminant (blue) and variational Bayesian logistic regression without ARD and with (red) or without hyper-priors (purple), and with ARD and hyper-priors on $\bm{w}$ (green).
    The figure illustrates the overly strong shrinkage of the methods without ARD (red, purple) due to the large number of uninformative dimensions.
    B) true class probability vs. estimated class probability for 200 observations of the training set.
    The colors are the same as in panel A, not showing Fisher's linear discriminant.
    The red shaded areas are areas in which mis-classification occurs.
    The strong coefficient shrinkage of methods without ARD is reflected by their predictive class probabilities being strongly biased towards 0.5.}
  \label{fig:logit_example_highdim}
\end{figure}

Evaluating the classification error on the training and test sets results in
\begin{lstlisting}
>> fprintf(['training set 0-1 loss: FLD    = %f, VB        = %f\n' ...
            '                       VBiter = %f, VB w/ ARD = %f\n'], ...
           mean(y_LD ~= y), ...
           mean(2 * (py_VB > 0.5) - 1 ~= y), ...
           mean(2 * (py_VB1 > 0.5) - 1 ~= y), ...
           mean(2 * (py_VB2 > 0.5) - 1 ~= y));
training set 0-1 loss: FLD    = 0.041500, VB        = 0.065500
                       VBiter = 0.112500, VB w/ ARD = 0.066000
>> fprintf(['test     set 0-1 loss: FLD    = %f, VB        = %f\n' ...
            '                       VBiter = %f, VB w/ ARD = %f\n'], ...
           mean(y_test_LD ~= y_test), ...
           mean(2 * (py_test_VB > 0.5) - 1 ~= y_test), ...
           mean(2 * (py_test_VB1 > 0.5) - 1 ~= y_test), ...
           mean(2 * (py_test_VB2 > 0.5) - 1 ~= y_test));
test     set 0-1 loss: FLD    = 0.282600, VB        = 0.260300
                       VBiter = 0.280200, VB w/ ARD = 0.203500
\end{lstlisting}
As can be seen, Fisher's linear discriminant over-fits the training set and thus features a higher test set loss than both variational methods with hyper-prior.
To understand the worse performance of the hyper-prior-free variational method (\code{VBiter}), it is instructive to investigates its coefficient estimates.
As seen in Fig.~\ref{fig:logit_example_highdim}A (purple diamonds), its prior on $\bm{w}$ with covariance $D^{-1} \bm{I}$, together with the large input dimensionality, $D=1000$, and low number of informative inputs $D_{\textrm{eff}} = 100$, caused the coefficient estimates to be close to zero, such that it severely underestimated these coefficients.
This is also reflected in its predictive class probabilities being close to 0.5 (Fig.~\ref{fig:logit_example_highdim}B, purple diamonds).
The variational method without ARD but with an adjustable degree of global shrinkage fares slightly better, but still under-estimates the informative coefficients due to the larger number of uninformative inputs (Fig~\ref{fig:logit_example_highdim}A, red circles).
This is again reflected in its predictive class probabilities being biased towards 0.5 (Fig.~\ref{fig:logit_example_highdim}B, red circles).
Finally, variational Bayesian logistic regression with ARD shows little signs of shrinkage of informative input dimensions (Fig.~\ref{fig:logit_example_highdim}A, green squares), which is also reflected in its wider spread of predictive class probabilities (Fig.~\ref{fig:logit_example_highdim}B, green squares).
Note that neither method performs perfectly, due to the task's considerable difficulty.

The code for this example is available in \code{vb\_logit\_example\_highdem.m}.

\subsubsection{Model selection by maximizing variational bound}

As for the linear case, the variational bound $\tilde{\varbound}(\VarProb, \bm{\xi})$ can be used as a proxy for the model log-evidence, $\ln \Prob(\data)$, to choose between models of different complexity.
We assume that a 1-dimensional scalar input is mapped into a polynomial of order $k$, which is in turn used to produce the class labels.
Given a set of observations of inputs and class labels, the task is to identify this order $k$ and train the associated classifier.
Assuming $k = 2$, we generate the data by
\begin{lstlisting}
>> D = 3;  N = 50;  Ds = 1:10;  x_range = [-5 5];
>> w = randn(D, 1);
>> x = x_range(1) + (x_range(2) - x_range(1)) * rand(N, 1);
>> x_test = linspace(x_range(1), x_range(2), 300)';
>> gen_X = @(x, d) bsxfun(@power, x, 0:(d-1));
>> X = gen_X(x, D);
>> py = 1 ./ (1 + exp(- X * w));
>> y = 2 * (rand(N, 1) < py) - 1;
>> py_test = 1 ./ (1 + exp(- gen_X(x_test, D) * w));
>> y_test = 2 * (rand(length(py_test), 1) < py_test) - 1;
\end{lstlisting}
In the above, \code{Ds} are the different orders to be tested, and the test data constitutes of 300 inputs spanning the range from -5 to 5.
The training data consists of 50 observations which are uniformly randomly distributed in the same range.

\begin{figure}
  \centering \includegraphics[width=9.5cm]{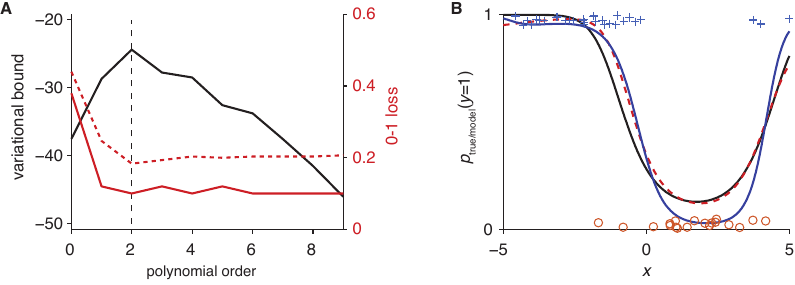}
  \caption{Variational bound and loss for models of different complexities, and model predictions.
    A) variational bound (black curve, left axis) and training/test set loss (solid/dashed red curve, right axis) for different assumed polynomial orders of the input.
    The black dashed line shows the true polynomial order.
    B) True class probability (black curve) vs. predicted class probability of variational Bayesian logistic regression (red dashed curve, using best-fit $k=2$ from A) and Fisher's logistic discriminant (blue curve, assuming $k = 5$) over scalar input $x$.
    The blue crosses and orange circles show the provided 50 training samples. They are scattered away from 0 and 1 for better visibility.}
  \label{fig:logit_example_modelsel}
\end{figure}

We find the variational bound and loss on training and test set for different orders of the polynomial by
\begin{lstlisting}
>> Ls = NaN(1, length(Ds));  pred_loss = NaN(length(Ds), 2);
>> for i = 1:length(Ds)
       [w, V, invV, ~, ~, Ls(i)] = vb_logit_fit(gen_X(x, Ds(i)), y);
       y_pred =  2 * (vb_logit_pred(gen_X(x, Ds(i)), w, V, invV) > 0.5) - 1;
       y_test_pred =  ...
           2 * (vb_logit_pred(gen_X(x_test, Ds(i)), w, V, invV) > 0.5) - 1;
       pred_loss(i, :) = [mean(y_pred ~= y) mean(y_test_pred ~= y_test)];
   end
>> [~, i] = max(Ls);
>> D_best = Ds(i)
D_best =

     3
\end{lstlisting}
Thus, for this particular training set, the method was able to correctly identify the underlying complexity.
Plotting the variational bound over model complexity shows a clear peak of the variational bound at this order (Fig.~\ref{fig:logit_example_modelsel}A).
Figure~\ref{fig:logit_example_modelsel}A also illustrates that, even though the training set error is equally low for polynomials of higher order (red, solid curve), Bayesian model selection takes the model complexity into account and thus selects a lower-order polynomial model.

For further testing, we train a Bayesian classifier with the identified polynomial order by
\begin{lstlisting}
>> X_VB = gen_X(x, D_best);
>> [w_VB, V_VB, invV_VB] = vb_logit_fit(X_VB, y);
>> py_VB = vb_logit_pred(X_VB, w_VB, V_VB, invV_VB);
>> py_test_VB = vb_logit_pred(gen_X(x_test, D_best), w_VB, V_VB, invV_VB);
\end{lstlisting}
Furthermore, we train Fisher's linear discriminant with assume polynomial order 5, by
\begin{lstlisting}
>> D_LD = 6;  y1 = y == 1;
>> X_LD = gen_X(x, D_LD);
>> w_LD = NaN(D_LD, 1);
>> w_LD(2:end) = (cov(X_LD(y1, 2:end)) + cov(X_LD(~y1, 2:end))) \ ...
                 (mean(X_LD(y1, 2:end))' - mean(X_LD(~y1, 2:end))');
>> w_LD(1) = - 0.5 * (mean(X_LD(y1, 2:end)) + ...
             mean(X_LD(~y1, 2:end))) * w_LD(2:end);
>> y_LD = 2 * (X_LD * w_LD > 0) - 1;
>> y_test_LD = 2 * (gen_X(x_test, D_LD) * w_LD > 0) - 1;
\end{lstlisting}
leading to training and test set losses of
\begin{lstlisting}
>> fprintf('Training set MSE, LD = %f, VB = %f\n', ...
           mean(y_LD ~= y), mean(2 * (py_VB > 0.5) - 1 ~= y));
Training set MSE, LD = 0.100000, VB = 0.100000
>> fprintf('Test     set MSE, LD = %f, VB = %f\n', ...
           mean(y_test_LD ~= y_test), ...
           mean(2 * (py_test_VB > 0.5) - 1 ~= y_test));
Test     set MSE, LD = 0.190000, VB = 0.183333
\end{lstlisting}
The slightly higher test-set loss for Fisher's linear discriminant is not surprising, given that its higher model complexity is bound to lead to over-fitting the training set data.
However, as illustrated in Fig.~\ref{fig:logit_example_modelsel}B, this over-fitting does not seem particularly severe in this particular case, as both the Bayesian estimate and the linear discriminant's estimate closely follow the true class probabilities.

The code for this example is available in \code{vb\_logit\_example\_modelsel.m}.

\section*{Acknowledgments}

I would like to thank Alexandre Pouget for his support, and Wei Ji Ma for comments on early versions of this manuscript.

\bibliographystyle{apalike}
\bibliography{refs}

\end{document}